\documentclass[10pt,twocolumn,letterpaper]{article}

\usepackage{mpi}
\usepackage{times}
\usepackage{epsfig}
\usepackage{graphicx}
\usepackage{amsmath}
\usepackage{amssymb}
\usepackage{multirow}
\usepackage{booktabs}
\usepackage{mathtools}
\usepackage[font={normalsize}]{caption}

\usepackage{commath}
\usepackage{hyperref}       %
\usepackage{placeins}
\usepackage{multirow}
\usepackage[acronym]{glossaries}
\usepackage{bm}
\newcommand{\vv}[1]{\bm{#1}}
\usepackage{array}
\newcolumntype{P}[1]{>{\centering\arraybackslash}p{#1}}
\usepackage[T1]{fontenc}
\providecommand{\e}[1]{\ensuremath{\times 10^{#1}}} %
\usepackage{graphicx}
\usepackage{subcaption}
\usepackage{cancel}

\newcommand{\br}[2]{\Tilde{\vv{r}}_{#1}(#2)}
\newcommand{\bri}[1]{\Tilde{\vv{r}}_i(#1)}
\newcommand{\ri}{\Tilde{\vv{r}}_i}
\newcommand{\f}{f}
\usepackage{placeins}
\usepackage{ifthen}
\usepackage{arydshln}

\makenoidxglossaries
\newacronym{sdf}{SDF}{Signed Distance Field}
\newacronym{sfm}{SfM}{Structure from Motion}
\newacronym{pdf}{PDF}{Probability Density Function}
\newacronym{cdf}{CDF}{Cumulative Distribution Function}
\newacronym{nrsfm}{NRSfM}{Non-Rigid Structure-from-Motion}
\newacronym{gan}{GAN}{Generative Adversarial Network}
\newacronym{mlp}{MLP}{Multi-Layer Perceptron}
\newacronym{icp}{ICP}{Iterative Closest Point}
\newacronym{mfof}{MFOF}{Multi-Frame Optical Flow}
\setkeys{glslink}{hyper=false} 

\newcommand{\suppfig}[3]{
\begin{figure}[ht!]
	\begin{center}
		\includegraphics[width=\linewidth]{#1}
	\end{center}
	\caption
	{
	    #2
	}
	\label{#3}
	\vspace{-12pt}
\end{figure}
}

\mpifinalcopy

\begin{document}

\title{Unbiased 4D: Monocular 4D Reconstruction with a Neural Deformation Model} 

\author{Erik C.M. Johnson\textsuperscript{1, 2}$\;\,$
Marc Habermann\textsuperscript{1}$\;\,$
Soshi Shimada\textsuperscript{1}$\;\,$
Vladislav Golyanik\textsuperscript{1}$\;\,$
Christian Theobalt\textsuperscript{1}
\vspace{5pt}
\\
\textsuperscript{1}Max Planck Institute for Informatics, SIC \ \ \ \ \ \ \ \ \textsuperscript{2}Saarland University, SIC
}

\twocolumn[{
\maketitle 
}]

\begin{abstract} 
Capturing general deforming scenes from monocular RGB video is crucial for many computer graphics and vision applications.
However, current approaches suffer from drawbacks such as struggling with large scene deformations, inaccurate shape completion or requiring 2D point tracks.
In contrast, our method, Ub4D, handles large deformations, performs shape completion in occluded regions, and can operate on monocular RGB videos directly by using differentiable volume rendering. 
This technique includes three new---in the context of non-rigid 3D reconstruction---components, i.e., 
1) A coordinate-based and implicit neural representation for non-rigid scenes, which in conjunction with differentiable volume rendering enables an unbiased reconstruction of dynamic scenes, 
2) a proof that extends the unbiased formulation of volume rendering to dynamic scenes, 
and 3) a novel dynamic scene flow loss, which enables the reconstruction of larger deformations by leveraging the coarse estimates of other methods.
Results on our new dataset, which will be made publicly available, demonstrate a clear improvement over the state of the art in terms of surface reconstruction accuracy and robustness to large deformations. 
\end{abstract}
\vspace{-0.5cm}
\section{Introduction}\label{sec:intro} 
\par 
Reconstructing the deforming 3D geometry of an object from image data is a long-standing and important problem in computer vision with many applications in the movie and game industries, as well as VR and AR.
Especially interesting and the subject of this work is the 4D reconstruction from a single RGB video, as this is the most intuitive and user-friendly capture setup.
Over the last decade, many monocular 4D reconstruction approaches have been proposed; they can be categorized into dense non-rigid structure from motion (NRSfM) methods, shape-from-template (SfT) approaches, and neural template-free approaches. 
\par
NRSfM methods~\cite{Bregler2000, Torresani2008,Garg2013, Ansari2017, Parashar2018,Kong2019,Novotny2019,sidhu2020neural} usually assume dense and coherent 2D point tracks connecting the frames of the video.
While accurate results can be obtained, it is usually hard to satisfy this assumption in real-world captures, limiting the use case in practice.
SfT methods~\cite{Salzmann2007, Ngo2015, yu2015direct,shimada2019ismo, FuentesJimenez2021,yu2015direct} assume an object template is given. 
While this provides a strong prior for this highly ill-posed task, initial reconstruction errors in the template can lead to tracking errors. 
Importantly, topological changes cannot be captured by such methods.
Last, template-free learning-based approaches have shown compelling results for category-specific (\textit{e.g.,} humans~\cite{saito2020pifuhd,saito19}) and general scenes with small deformations~\cite{Tretschk2020, pumarola2020dnerf}. 
However, generalization beyond categories and accurate reconstruction of an explicit geometry remains a challenge.

\begin{figure}
    \includegraphics[width=\linewidth]{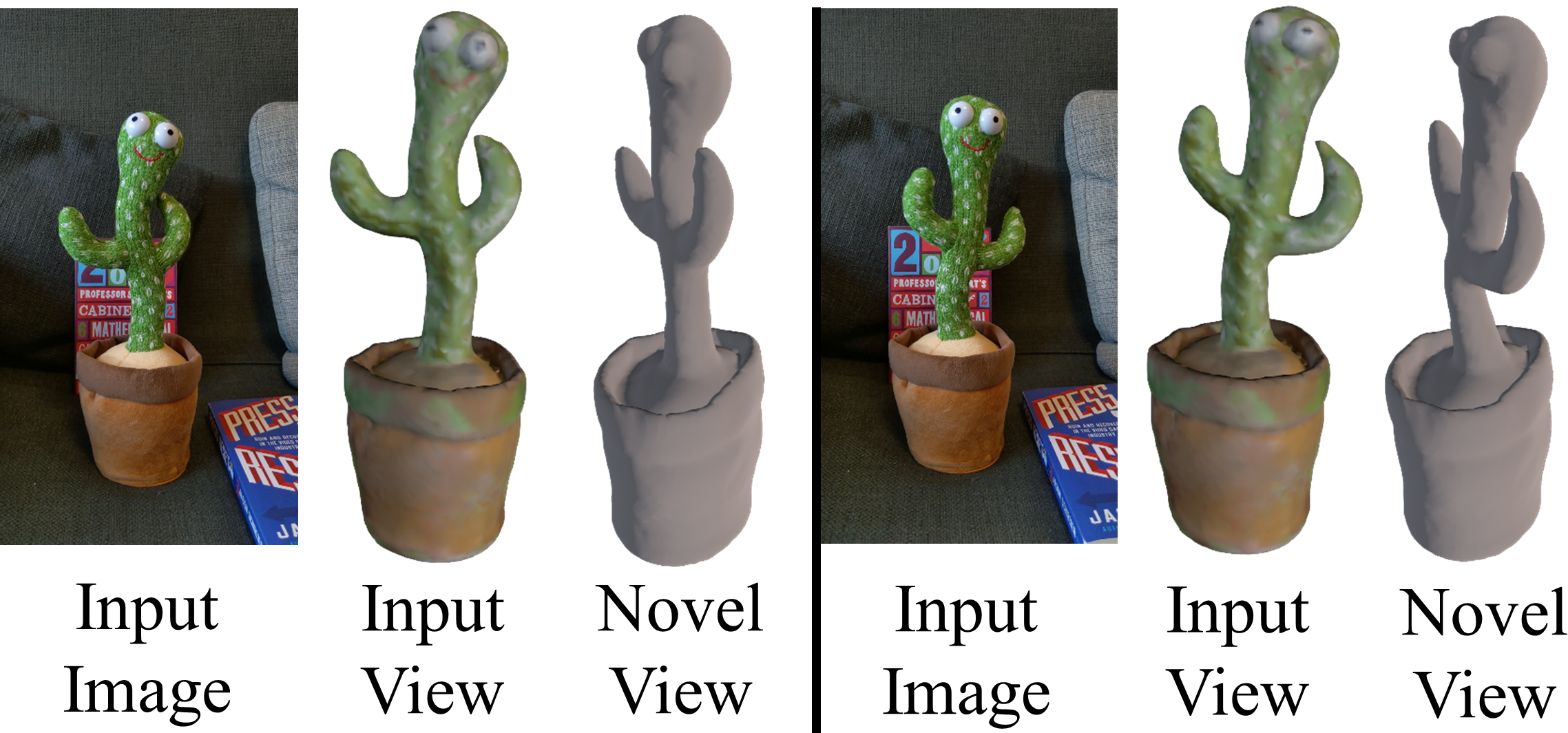}
    \caption
    {
        We present a new method for 4D reconstruction of dynamic scenes using a single RGB video. 
    	In contrast to previous work, our method completes the object as it is observed from different view angles, can handle small- and large-scale deformations of arbitrary objects due to our separation of non-rigid deformations using a canonical space, our unbiased volume rendering formulation, and an optional scene flow loss. 
	}
	\label{fig:teaser} 
	\vspace{-0.5cm}
\end{figure}

%
%
\par 
To this end, we propose Unbiased 4D (Ub4D), \textit{i.e.,} a novel method for the 4D reconstruction of a deforming object given a single RGB video of the object; see Figure~\ref{fig:teaser}. 
In contrast to previous method classes such as non-rigid structure-from-motion and shape-from-template, Ub4D completes the shape as it is being observed from other view angles.
Using a signed distance field (SDF) network, we represent the object of interest as an implicit SDF in canonical space. 
In order to obtain the deformed per-frame geometry, we propose a bending network, which deforms the current frame into a shared canonical space.
To supervise the SDF and bending network, we impose an unbiased volume rendering loss, which extends prior work~\cite{wang2021neus} to dynamic scenes.
Notably, we also prove the correctness of our formulation.
In particular, we compare the rendered images and object segmentation masks with the ground truth images and masks.
Similar to previous works~\cite{Tretschk2020}, this formulation alone still struggles with larger scene deformations.
Thus, when the scene contains large scene deformations, we propose a scene flow loss, which attaches free space to a set of tracked 3D points in order to guide the scene deformations predicted by the bending network. 
%
%
In summary, our primary technical  contributions are as follows: 
\begin{itemize}\itemsep0em 
\setlength{\parskip}{2pt} 
\item{Ub4D, \textit{i.e.,} a new approach for dense 4D reconstruction from monocular image sequences based on an implicit surface representation and a dynamic bending network (Sec.~\ref{sec:approach}).} 
\item{Extending the unbiased formulation of volume  rendering~\cite{wang2021neus} to general deforming scenes (Sec.~\ref{sec:render}).} 

\item{A new scene flow loss leveraging coarse geometric proxies (dense and sparse), which further increases the robustness to large-scale scene deformations (Sec.~\ref{sec:scene_flow_loss}).} 
\item A new synthetic 
benchmark dataset for general and large-scale deforming scenes (Sec.~\ref{sec:results}). 
\end{itemize} 
We demonstrate that our method outperforms the previous  state of the art in terms of accuracy and robustness to large scale scene deformations (Secs.~\ref{sec:quan_compare}-\ref{sec:qual_results}). 
Our code and the new dataset will be made publicly available. 
\section{Related Work}\label{sec:related_work} 
 
Several method classes for 3D reconstruction of non-rigidly deforming surfaces from monocular images are known. %
They differ in the assumptions they make about the available priors and types of motions and deformations. 

\noindent\textbf{Non-Rigid Structure from Motion (NRSfM)} 
operates on point tracks over the input monocular views \cite{Bregler2000, Torresani2008}. 
It factorizes them into camera poses and deformable (per-frame) geometry of observed surfaces. 
Assuming that accurate point tracks can be obtained is a restrictive assumption. 
If points of the input views are tracked densely, NRSfM can then even be used to obtain dense surfaces \cite{Garg2013, Ansari2017, Parashar2018}. 
Both neural NRSfM methods for the sparse \cite{Kong2019,Novotny2019,Wang2021} and dense  \cite{sidhu2020neural} cases were recently proposed in the literature. 
Deep NRSfM \cite{Kong2019,Novotny2019,Wang2021,sidhu2020neural} is related to NRSfM in that it lifts 2D input points in 3D and does not rely on 3D supervision. 
Ub4D is similar to NRSfM in that 1) it has the least number of assumptions (no training datasets, no 3D priors) and 2) requires camera or object movement while recording the scene. 
It differs from NRSfM in that it operates directly on images with no need for 2D correspondences. 
\noindent\textbf{Shape from Template (SfT).} 
This class of techniques assumes a 3D shape prior called a \textit{template}. 
SfT is then posed as the problem of tracking and deforming the template so that the new states plausibly reproject to the input images \cite{Salzmann2007, Ngo2015, yu2015direct,NRST_GCPR2018}.
While some approaches have demonstrated accurate results, even for larger deformations, they come at the cost of being category-specific (\textit{e.g.,} they only work for humans~\cite{habermann2019,Bogo:ECCV:2016}). 
Further, the assumption of a known 3D template is limiting when dealing with unknown objects. 
Moreover, obtaining the template usually requires a separate step, which can be difficult. 
$\phi$-SfT \cite{Kairanda2022} explains 2D observations through physics-based simulation of the deformation process. 
In contrast to them, we do not model physics laws explicitly. 
Moreover, we target a different class of non-rigid objects (thin surfaces \cite{Kairanda2022} \textit{vs} articulated objects). 
Deep SfT or direct surface regression methods assume multiple states  available for training \cite{shimada2019ismo, FuentesJimenez2021}. 
Our approach differs from SfT in that it only requires 2D images as input. 
Nonetheless, it can benefit from a subset of frames observing static scene states to initialise the canonical volume. %
Note that---in contrast to SfT techniques---observing a scene under rigidity assumption \cite{yu2015direct} or having a template in advance from elsewhere is not a strict requirement for us. 

\noindent\textbf{Monocular 3D Mesh Reconstruction.} 
3D mesh reconstruction methods deform an initial mesh to match image observations \cite{Wang2018Pixel2Mesh,Kanazawa2018,VMR2020,yang2021lasr}. 
They are exclusively neural techniques, usually trained using 2D 
image collections. 
One of their limitations 
is that large image sets are not available for all object  categories (\textit{e.g.,} consider rarely observed  biological species).
Moreover, the methods, which do not require 2D image priors, might capture coarse articulations but fail to reconstruct fine surface details \cite{Kanazawa2018,yang2021lasr}. 
Starting from a sphere mesh is a restricting assumption. 
Even though many watertight meshes are, in theory, topologically equivalent to a sphere, a practical attempt to guide sphere  deformations by image cues can converge to local minima. 
\noindent\textbf{Free Viewpoint Video and Neural Surface Extraction.} 
Coordinate-based volumetric neural representations learned from 2D observations, such as NeRF \cite{mildenhall2020nerf}, can be used to render high-quality novel views of rigid  \cite{yu2021plenoctrees, Chibane2021,liu2020neural} and non-rigid scenes  \cite{Tretschk2020, Peng_2021_CVPR, liu2021neural,peng2021animatable,park2020nerfies, pumarola2020dnerf,Li2020NeuralSF,xian2020spacetime}. 
While they have shown impressive 
results, the volumetric representation they use lacks surface constraints so that it is difficult to extract high-quality surfaces from the learned representation.
Some works \cite{Oechsle2021, wang2021neus,yariv2021volume, yariv2020multiview} propose to represent 3D scenes as a neural SDF and use volume rendering to learn the representation. 
\begin{figure*}[t!]
	\begin{center}
		\includegraphics[width=\linewidth]{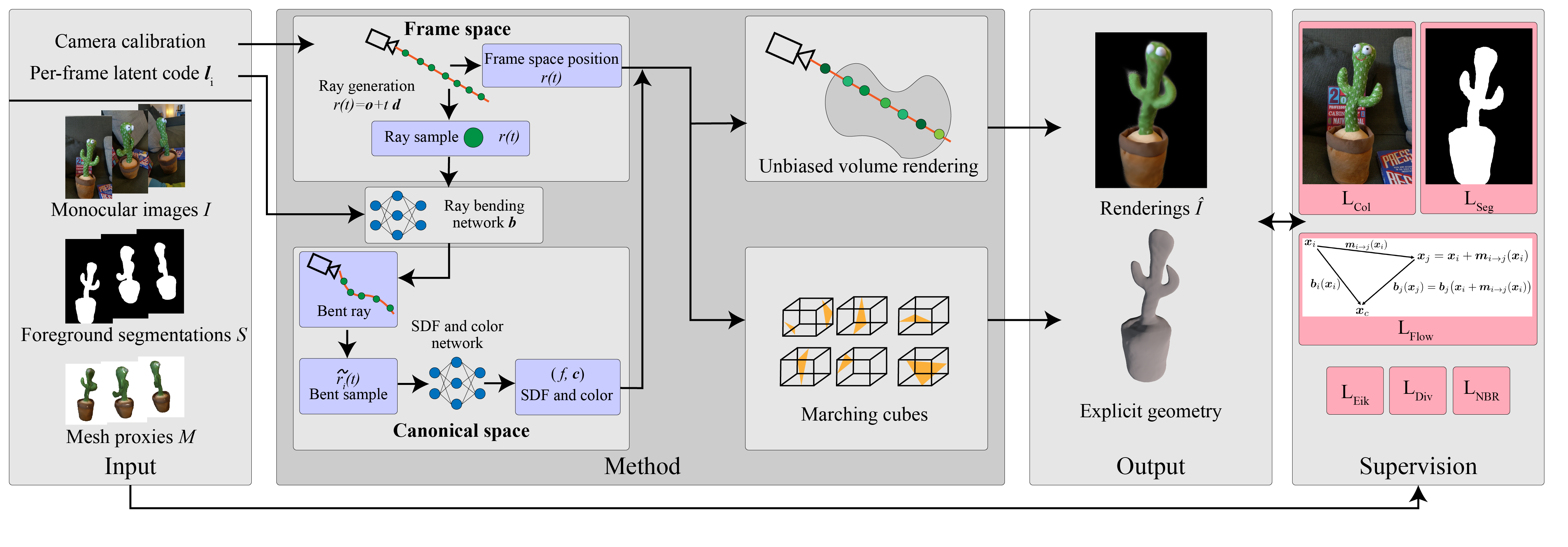}
	\end{center}
    \vspace{-0.8cm}
	\caption
	{
    	Ub4D takes as input a sequence of images recorded with a single RGB camera and foreground masks. 
    	Each frame is also equipped with a learnable latent code, 
    	and our method learns a canonical and colored SDF representing the static scene. 
    	Our bending network maps the frame space to canonical space and volume rendering and marching cubes can produce per-frame renderings and geometries. 
    	We weakly supervise our scene representation with image-based losses and spatiotemporal priors including our novel scene flow loss. 
	}
    \vspace{-0.4cm}
	\label{fig:overview}
\end{figure*}
We are inspired by the recent progress in neural volumetric representations learned without 3D supervision. 
Even though our goal is not novel view rendering and  editing, we show that a NeRF-inspired component can be useful for monocular non-rigid 3D reconstruction. 
Moreover, surface extraction methods \cite{Oechsle2021, wang2021neus,yariv2021volume, yariv2020multiview} have focused on rigid objects so far. 
We demonstrate that the problem we are interested in, \textit{i.e.,} monocular \textit{non-rigid} 3D reconstruction, significantly benefits from advances in another, distantly related research direction. 
%
 
%
%
\section{Method}\label{sec:approach}
The goal of Ub4D is to reconstruct the dense and deforming surface of an object from a single RGB video.
Therefore, our method takes as input the monocular image sequence $\{I_i, S_i: i \in [1, N_f]\}$ of the segmented object consisting of $N_f$ RGB images $I_i$ and respective segmentation masks $S_i$.
We assume the extrinsic and intrinsic camera parameters are known.
Optionally, corresponding per-frame coarse geometric proxies with $N_v$ vertices can be provided $\{M_i: i \in [1, N_f]\}$, where $M_i = \{\vv{v}_i^{(k)}: k \in [1, N_v]\}$ and $\vv{v}_i^{(k)}$ denotes vertex $k$ of the mesh in frame $i$.
Note that we only use corresponding vertices, i.e. no connectivity information, which allows the use of sparse point sets (e.g. skeleton) as the geometric proxy.
Given these inputs, Ub4D outputs an explicit geometry for every frame; see Fig.~\ref{fig:overview}. 
%
%
\subsection{Non-Rigidity Model} \label{sec:nr_model}
We model temporal non-rigid deformations as a vector field projecting points from the frame space into a canonical space.
One can conceptualize this by considering it as a bending of the straight rays originating from the camera.
Given a straight ray with an origin $\vv{o} \in \mathbb{R}^3$ and a viewing direction $\vv{d} \in \mathbb{R}^3$ as $\vv{r}(t) = \vv{o} + t \vv{d}$, we bend this ray with a frame-specific bending network $\vv{b}_i: \mathbb{R}^3 \rightarrow \mathbb{R}^3$ as $\Tilde{\vv{r}}_i(t) = \vv{r}(t) + \vv{b}_i (\vv{r}(t))$ where $i$ denotes the frame.
This bent ray is a directed parametric path in $\mathbb{R}^3$ like the straight ray, but, where the derivative of the straight ray is constant (\textit{i.e.} $\od{\vv{r}(t)}{t} = \vv{d}$), the bent ray has an instantaneous direction at each point along it of $\od{\Tilde{\vv{r}}_i(t)}{t} = \vv{d} + \frac{\partial \vv{b}_i}{\partial \vv{r}(t)} \vv{d}$.
We desire that this bending network transforms points from frame space into a single canonical representation of the object shared by all frames of the input.
\par
While this discussion presents the bending network as a per-frame vector field throughout $\mathbb{R}^3$, it is implemented using a per-frame latent code $\vv{l}_i \in \mathbb{R}^{64}$.
This latent code is given as input along with a point in space to \ifglsused{mlp}{an}{a} \gls{mlp} $\vv{b}: (\mathbb{R}^3, \mathbb{R}^{64}) \rightarrow \mathbb{R}^3$ and the latent codes are optimized during training. 
Importantly, the latent code $\vv{l}_i$ passed to the bending network is the \textit{only} frame-specific element in our method and no other network receives it. 
\par
This is similar to the non-rigidity model employed in NR-NeRF~\cite{Tretschk2020}. 
However, we propose a different regularization to enable the modeling of larger deformations, which also removes the need to learn a rigidity score throughout the scene.
Whereas NR-NeRF~\cite{Tretschk2020} penalizes the bending network output for its \textit{absolute} length, we instead enforce that the deformation of the current frame is similar to that of the neighboring frames.
This assumes that neighboring frames represent similar object states, which is a more reasonable assumption for dynamic scenes compared to the absolute amount of deformation.
More specifically, for $N_s$ samples along a straight ray $\vv{r}$, we penalize the bending network as:
%
%
\begin{equation} 
\small 
    L_{\text{NBR}} = \frac{1}{N_s} \sum_{z=1}^{N_s} \sum_{j \in \mathcal{N}(i)} \omega_i^{(z)}\, ||\vv{b}_i (\vv{r}(t^{(z)})) - \vv{b}_j (\vv{r}(t^{(z)})) ||_2^2, 
\end{equation}
%
where $\omega_i^{(z)}$ is the visibility weight at sample $z$ along the bent ray (see Sec.~\ref{sec:render}) and $\mathcal{N}(i)$ are the neighbours of frame $i$.
We also penalize the divergence of the bending network as:
%
%
\begin{align} \label{eqn:divergence_loss}
    L_{\text{DIV}} &= \frac{1}{N_s} \sum_{z=1}^{N_s} \omega_i^{(z)}\, | \nabla \cdot  \vv{b}_i (\vv{r}(t^{(z)}))|^2, 
\end{align} 
%
%
where we use the unbiased, approximated divergence as  presented in Tretschk \textit{et al.}~\cite{Tretschk2020}. 
%
%
\subsection{Rendering Method}\label{sec:render}
Recent studies on \textit{static} scene reconstruction have demonstrated that volume rendering enables more stable training compared to surface rendering~\cite{yariv2020multiview, wang2021neus, yariv2021volume}. 
Therefore, we extend the volume rendering method proposed in NeuS~\cite{wang2021neus} to \textit{dynamic} scenes. 
The full proof of unbiasedness is in the supplement.
Let $f: \mathbb{R}^3 \rightarrow \mathbb{R}$ be \ifglsused{sdf}{an}{a} \gls{sdf} modeled by \ifglsused{mlp}{an}{a} \gls{mlp} that takes as input sampled points $\Tilde{\vv{r}}_i (t)$ on the bent ray. 
Then, NeuS~\cite{wang2021neus} shows that we can compute the opaque density as:
\begin{align} \label{eqn:density}
    \rho_i(\Tilde{\vv{r}}_i (t)) &= \text{max} \Bigg\{ \frac{-\od{\Phi_s}{t}\big( f(\Tilde{\vv{r}}_i(t)) \big)}{\Phi_s \big( f(\Tilde{\vv{r}}_i(t)) \big)}, 0 \Bigg\},
\end{align}
%
%
where $\Phi_s$ is the logistic \gls{cdf} with standard deviation $s^{-1}$.
This is in contrast to the original formulation operating on unbent ray sample points rather then bent ones. 
To calculate the color of each camera ray, we employ a hierarchical sampling with $N_s$ samples in total (samples in coarse and fine stages) along the \textit{bent} ray $\{ \Tilde{\vv{r}}_i (t^{(z)}): z \in \mathbb{Z}, z \in [1, N_s] \}$ where $t^{(z)} < t^{(z+1)}, \forall z$.
Then, the color of the ray can be computed as:
%
%
\begin{align}
\hspace{-5pt}
    \hat{I}(\Tilde{\vv{r}}_i) = \sum_{z=1}^{N_s-1} \omega_i^{(z)}\, \vv{c} \Big( \Tilde{\vv{r}}_i\big(t^{(z)}\big), \Tilde{\vv{r}}_i\big(t^{(z+1)}\big) - \Tilde{\vv{r}}_i\big(t^{(z)}\big) \Big),
\end{align}
%
%
where $\vv{c}(\cdot)$ is a color function modeled by an MLP, which takes as input the point position $\Tilde{\vv{r}}_i(t)$ and the viewing direction of the ray \textit{at that point}, which is approximated with a forward difference.
The weight $\omega_i^{(z)}$ is occlusion-aware and \textit{unbiased} with respect to the object's surface (see our supplementary material for the proof), which is formulated based on the opaque density $\rho_i(\Tilde{\vv{r}}_i (t))$ from Equation~\eqref{eqn:density} as follows: 
%
%
\begin{equation} 
\omega_i^{(z)} = T_i^{(z)}\, \alpha_i^{(z)}, \;\text{with}\: 
T_i^{(z)} = \prod_{\zeta=1}^{z-1} (1 - \alpha_i^{(\zeta)}),\;\text{and}
\end{equation}
%
%
%
%
%
{\footnotesize
\begin{align}
    \alpha_i^{(z)} = \text{max} \Bigg\{ \frac{\Phi_s\Big(f\big(\Tilde{\vv{r}}_i(t^{(z)})\big)\Big) - \Phi_s\Big(f\big(\Tilde{\vv{r}}_i(t^{(z+1)})\big)\Big)}{\Phi_s\Big(f\big(\Tilde{\vv{r}}_i(t^{(z)})\big)\Big)}, 0 \Bigg\},
\end{align}
}
%
%
\begin{equation}
\alpha_i^{(z)}=1-\exp\left(-\int_{t^{(z)}}^{t^{(z+1)}}\rho(t){\rm d}t\right).
\label{discrete_alpha_0}
\end{equation}
%
%
Importantly, the discrete opacity $\alpha_i^{(z)}$  derivation 
\cite{wang2021neus} still applies in the case of a bent ray as replacing the constant viewing direction with $\od{\Tilde{\vv{r}}_i(t)}{t}$ does not affect the analysis. 
\par 
In addition to the color, we can determine if a ray intersects the object by computing the sum of the weights:
%
%
\begin{align} \label{eqn:discrete_seg}
    \hat{S}(\Tilde{\vv{r}}_i) &= \sum_{z=1}^{N_s-1} \omega_i^{(z)},
\end{align}
%
%
where $\hat{S}$ approaches $1$ for a ray intersecting the object and otherwise $\hat{S}$ approaches $0$.
%
%

\noindent\textbf{Supervision.} \label{sec:recon_losses} 
We supervise the dynamic scene representation by
$\ell_1$-distance between the color of each bent ray $\Tilde{\vv{r}}_i^{(p)}$ and the corresponding ground-truth color $I_i^{(p)}$ of pixel $p$: 
%
%
\begin{align} \label{eqn:colour_loss}
    L_{\text{COL}} &= \frac{1}{N_p} \sum_{p=1}^{N_p} \Big| \hat{I} \big( \Tilde{\vv{r}}_i^{(p)} \big) - I_i^{(p)} \Big|, 
\end{align}
%
%
where $N_p$ is the number of pixels sampled from frame $i$.
To more explicitly ensure that our approach solely focuses on reconstructing the foreground object, we also define a segmentation loss $L_{\text{SEG}}$ as the binary cross entropy between the estimated segmentation $\hat{S} \big( \Tilde{\vv{r}}_i^{(p)} \big)$ and the ground truth object segmentation $S_i^{(p)}$.
Finally, we enforce $f$ to be an SDF with the Eikonal loss defined as follows:
%
%
\begin{align} \label{eqn:eikonal_loss}
    L_{\text{EIK}} &= \frac{1}{N_p N_s} \sum_{p=1}^{N_p} \sum_{z=1}^{N_s} (|\nabla f(\Tilde{\vv{r}}_i^{(p)} (t^{(z)}))| - 1)^2. 
\end{align}
\subsection{Optional Scene Flow Loss} \label{sec:scene_flow_loss}
So far, very large scene deformations remain a challenge for Ub4D since it can create erroneous multiple geometries in the canonical space to best explain the monocular observations.
This is particularly noticeable for scenes containing large translations like \textit{RootTrans} (see Figure~\ref{fig:vs_nrnerf_ablation}-(b)).
To resolve this, we accept an additional input in the form of a coarse and coherent per-frame geometric proxy.
From these coarse 3D correspondences, we can compute an estimate of the scene flow, which can then be used to regularize the bending network.
This greatly reduces the effect of duplicate geometries in the canonical space.
\textit{Note that the scope of this work is not how such coarse proxies are obtained, rather we focus on how these enable Ub4D to densely track larger scene deformations.} 
\par
Consider a function $\vv{m}_{i \rightarrow j}: \mathbb{R}^3 \rightarrow \mathbb{R}^3$ that returns the scene flow estimate at a point from frame $i$ to $j$.
The scene flow allows us to transform points from a frame $i$ into any other frame $j$ as $\vv{x}_j = \vv{x}_i + \vv{m}_{i \rightarrow j} (\vv{x}_i)$.
Given that the bending network projects a point $\vv{x}_i$ in frame space into canonical space resulting in the point $\vv{x}_c$, it follows:
\begin{align} \label{eqn:sceneflow_prior}
\begin{split}
    \vv{x}_c &= \vv{x}_i + \vv{b}_i (\vv{x}_i) = \vv{x}_j + \vv{b}_j (\vv{x}_j) \\
    &= \vv{x}_i + \vv{m}_{i \rightarrow j} (\vv{x}_i) + \vv{b}_{j} \big(\vv{x}_i + \vv{m}_{i \rightarrow j} (\vv{x}_i)\big).
\end{split}
\end{align}
Intuitively, this means that a point in frame $i$ and its corresponding point in frame $j$ determined through the scene flow $\vv{m}_{i \rightarrow j} (\vv{x}_i)$ should be mapped to the same point $\vv{x}_c$ in canonical space by the bending network (see Fig.~\ref{fig:sceneflow_prior}).
We can then formulate it as a loss for a set $\mathcal{X}$ of sampled points:
{\small 
\begin{align}
    L_{\text{FLO}} = \frac{1}{|\mathcal{X}|} \sum_{\vv{x} \in \mathcal{X}} || \vv{m}_{i \rightarrow j} (\vv{x}) + \vv{b}_j \big(\vv{x} + \vv{m}_{i \rightarrow j} (\vv{x})\big) - \vv{b}_i (\vv{x}) ||_2^2.
\end{align}
}
\begin{figure}[t]
	\begin{center}
		\includegraphics[width=0.6\linewidth]{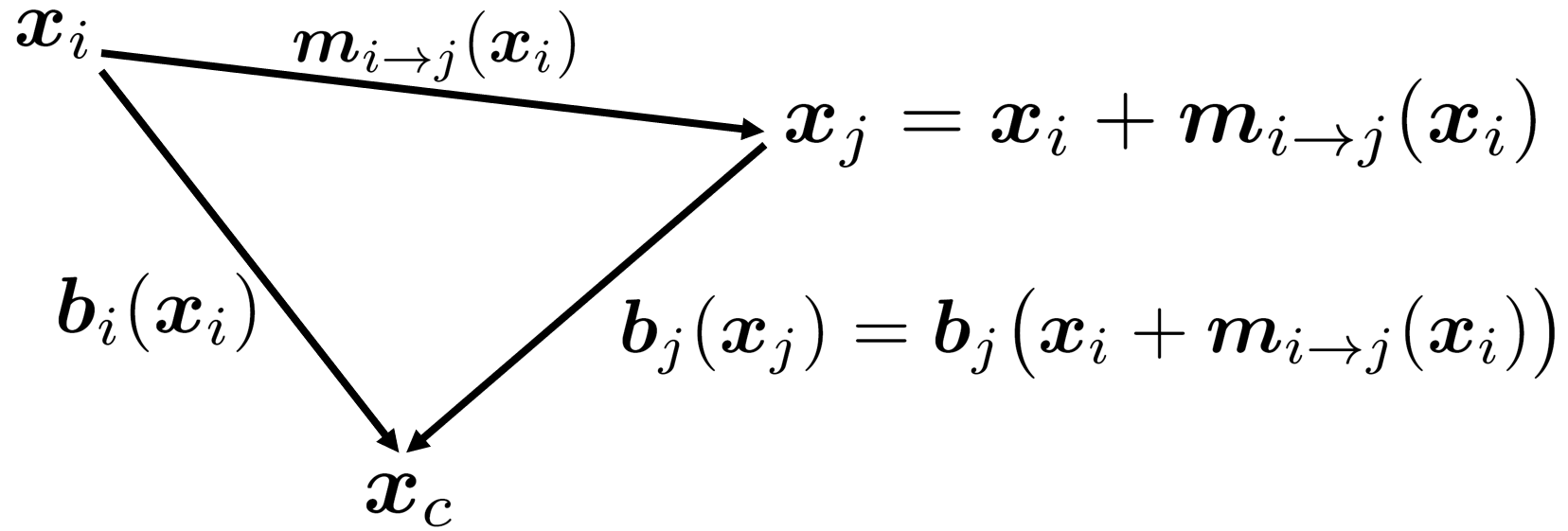}
	\end{center} 
    \vspace{-0.25cm}
	\caption 
	{
	    Graphical depiction of the relationship between the scene flow from frame $i$ to $j$ (\textit{i.e.,} $\vv{m}_{i \rightarrow j} (\vv{x}_i)$) and the bending network projecting both points to the same canonical position $\vv{x}_c$.
	}
    \vspace{-0.25cm}
	\label{fig:sceneflow_prior}
\end{figure}
\par
The scene flow from the proxies can only be exactly computed at the proxy vertices $\vv{v}_i^{(k)}$ but
our implicit surface representation can be evaluated at any point in 3D space.
Thus, we extrapolate this scene flow to any point in $\mathbb{R}^3$ with a convex combination over the vertices using a kernel function depending on the distance to the vertices inspired by the spatial weighting approach in bilateral filtering~\cite{tomasi1998bilateral}:
\begin{align}
    \vv{m}'_{i \rightarrow j} (\vv{x}) = \frac{\sum_{k=1}^{N_v} w_{\lambda_1}\big(||\vv{x} - \vv{v}_i^{(k)}||_2\big)\, \big(\vv{v}_j^{(k)} - \vv{v}_i^{(k)}\big)}{\sum_{k=1}^{N_v} w_{\lambda_1}\big(||\vv{x} - \vv{v}_i^{(k)}||_2\big)}, 
\end{align}
where $w_{\lambda_1}(x) = e^{-\lambda_1 x^2}$ is a kernel function with a parameter $\lambda_1$  affecting the weighting of vertex flow estimates.
Additionally, we add an attenuation term, so that the scene flow falls off as the distance to the nearest vertex increases:
\begin{align} \label{eqn:sceneflow_falloff}
    \vv{m}_{i \rightarrow j} (\vv{x}) &= w_{\lambda_2}\Big(\min_{k=1}^{N_v} ||\vv{x} - \vv{v}_i^{(k)}||_2\Big)\, \vv{m}'_{i \rightarrow j} (\vv{x})
\end{align}
where $w_{\lambda_2}(x) = e^{-\lambda_2 x^2}$ is a kernel function with $\lambda_2$ as a scale parameter defining the extent of the kernel.
\subsection{Surface Extraction} \label{sec:surface}
To convert our deforming and implicit scene representation into an explicit geometry, we use the Marching Cubes algorithm~\cite{lorensen1987marching}  with a threshold of 0.
For points sampled in frame $i$, we transform them from the frame space into the canonical space, \textit{i.e.,} $\vv{x}_c = \vv{x}_i + \vv{b}_i(\vv{x}_i)$,
where $\vv{x}_i$ is a point sampled for marching cubes and $\vv{x}_c$ is the canonical space point at which we then evaluate the SDF.
We restrict the selection of frame march points to the camera frustum of the given frame since any space not seen in that frame is unconstrained by our reconstruction losses and may contain aberrant geometry.
\section{Experimental Results} \label{sec:results} 
We first introduce the datasets we are using for evaluation as well as the evaluation metrics. 
Next, we 
compare our method to previous works on monocular 4D scene reconstruction (Sec.~\ref{sec:quan_compare}). 
We validate the design choice of using an SDF network rather than a density network (Sec.~\ref{sec:volume_rep}). 
Finally, we ablate important design choices (Sec.~\ref{sec:ablations}) and show more qualitative results on real world data (Sec.~\ref{sec:qual_results}). 
All tests were performed using a single NVIDIA Quadro RTX 8000 with $48$~GB RAM. 
%
%
\par \noindent \textbf{Dataset.} %
We aim at reconstructing the \textit{full} deforming geometry and, thus, the monocular capture requires sufficient camera motion around the dynamic object to observe every part at least once.
However, we found that existing datasets either capture static scenes with a circulating camera path around the object or dynamic scenes with very limited camera motion.
Therefore, we create our own dataset of dynamic objects with sufficient camera motion, which contains synthetic scenes for quantitative evaluations and real scenes for qualitative results.
%
%

%
For the synthetic evaluation, we create two scenes in Blender~\cite{Hess2010} 
showing a deforming cactus, referred to as \textit{Cactus}, and a moving human, referred to as \textit{RootTrans}. 
Each of the scenes has an image resolution of $1024{\times}1024$ and is $150$ frames long. 
We define a moving camera viewing the dynamic object and provide the camera parameters as input to our method.
To generate the optional proxy geometries, which are used in our proposed scene flow loss for capturing large deformation, we leverage a human capture method~\cite{SMPL-X:2019} for the \textit{RootTrans} sequence (further details are included in our supplementary material).
For the \textit{Cactus} sequence, we use a highly downsampled version of the ground-truth geometry as a coarse proxy.
A visualization of the proxies is shown in Figure~\ref{fig:qual_comparison}.
%
%
%
\par
For the evaluation of our method on real data, we capture two sequences: one of a moving human, called the \textit{Humanoid} sequence, and one of a deforming cactus toy, called \textit{RealCactus}.
We capture these sequences at resolutions of $960{\times}1280$ and $1080{\times}1920$, respectively, with a mobile phone camera.
Again each sequence contains around 150 frames.
To obtain the camera parameters, we use the rigid \gls{sfm} software COLMAP~\cite{schoenberger2016sfm, schoenberger2016mvs}.
As with the \textit{RootTrans} synthetic sequence, we generate proxy geometries for the \textit{Humanoid} sequence using the same human capture method~\cite{SMPL-X:2019}.
However, unlike the \textit{RootTrans} sequence, we only input the \textit{joint positions} (i.e. 12 vertices) as the proxy, rather than the full posed SMPL-X~\cite{SMPL-X:2019} model.
This shows that our proxy need not include any information about the location of the surface.
To demonstrate the \textit{optional} nature of the scene flow loss, the \textit{RealCactus} sequence does not use any proxy geometry at all.
For the foreground masks, we manually labeled a few frames and then trained a segmentation network, based on the UNet architecture~\cite{unet}, on those labeled frames, which then provides masks for all frames in a semi-automated fashion.
%
%
\par %
For additional evaluation of our method, we leverage the \textit{Lego} object\footnote{Released under \href{https://creativecommons.org/licenses/by/3.0/legalcode}{CC-BY-3.0} and modifications are made. Originally created by Blend Swap user Heinzelnisse.}
made available by Mildenhall \textit{et al.}~\cite{mildenhall2020nerf}, which we animate over time by lifting the boom and tilting the bucket (see Fig.~\ref{fig:vs_nrnerf_ablation}-(a)), to obtain a dynamic scene.
Further, we defined a monocular camera path for 150 frames, rendered monocular images and masks at a resolution of $800{\times}800$, and extracted the camera extrinsics and intrinsics.
This scene does not use 3D proxy for our method. 
%
%
\par \noindent \textbf{Evaluation Metrics.}
To quantitatively %
compare Ub4D to the state-of-the-art methods, we compute the Chamfer distance (CD) between the estimated geometry and the ground-truth geometry. 
For a more fair comparison with previous monocular non-rigid 3D reconstruction methods, 
we break the CD down into its two components: measuring from estimate to ground-truth ($\textit{E2G}$) and from ground-truth to estimate ($\textit{G2E}$). 
The reported numbers are in relative distance units since synthetic scenes do not have an interpretable physical scale. 
NR-NeRF~\cite{Tretschk2020} and D-NeRF~\cite{pumarola2020dnerf} are given the same camera parameters used by our method.
Some methods either assume a fixed camera~\cite{yu2015direct} or predict the camera poses~\cite{yang2021lasr, sidhu2020neural, yang2021viser}.
For these cases, we apply ICP~\cite{besl1992icp} to rigidly align their meshes with the ground truth before computing metrics.

\begin{table}[t] 
        \begin{center}
    		\begin{tabular}{|c|c|c|c|c|}
    			\hline
    			& \textit{Method} & \textit{CD} ($\downarrow$) & \textit{E2G} ($\downarrow$) & \textit{G2E} ($\downarrow$) \\
    			\hline
    			{\parbox[t]{2mm}{\multirow{8}{*}{\rotatebox[origin=c]{90}{\textit{Cactus}}}}} & D-NeRF \cite{pumarola2020dnerf} & [113.38] & - & 7.99 \\
    			& NR-NeRF \cite{Tretschk2020} & [96.96] & - & 9.89 \\
    			& LASR \cite{yang2021lasr} & 20.23 & 12.09 & 8.14 \\
    			& ViSER \cite{yang2021viser} & 14.34 & 7.93 & 6.41 \\
    			& N-NRSfM \cite{sidhu2020neural} & [102.00] & 5.74 & - \\
    			& DDD \cite{yu2015direct} & 34.71 & 6.98 & 27.72 \\
    			& \textbf{Ub4D (ours)} & \textbf{3.06} & 2.42 & 0.64 \\
    			\cdashline{2-5}
    			& Ub4D after ICP & 2.71 & 2.24 & 0.46 \\
    			\hline
    			{\parbox[t]{2mm}{\multirow{8}{*}{\rotatebox[origin=c]{90}{\textit{RootTrans}}}}} & D-NeRF \cite{pumarola2020dnerf} & [23.50] & - & 8.43 \\
    			& NR-NeRF \cite{Tretschk2020} & [1.94] & - & 0.33 \\
    			& LASR \cite{yang2021lasr} & 0.39 & 0.08 & 0.31 \\
    			& ViSER \cite{yang2021viser} & 0.37 & 0.20 & 0.17 \\
    			& N-NRSfM \cite{sidhu2020neural} & [0.38] & 0.09 & - \\
    			& DDD \cite{yu2015direct} & 0.26 & 0.10 & 0.16 \\
    			& \textbf{Ub4D (ours)} & \textbf{0.23} & 0.14 & 0.09 \\
    			\cdashline{2-5}
    			& Ub4D after ICP & 0.03 & 0.02 & 0.02 \\
    			\hline
    		\end{tabular}
    	\end{center}
      
    	 \vspace{-0.5cm}
    	\caption { \label{tab:quan_comparison}
        \textbf{Quantitative comparison to previous work.} We include the unidirectional breakdown of the CD as well, since some methods produce geometries for which the CD is unfair as a metric (these are denoted by placing the CD in square brackets and omitting the unfair metric). See our supplement for further discussion.
        }
        \vspace{-0.5cm}
\end{table}
	\hfill 
\begin{table}%
        \begin{center}
    		\begin{tabular}{|c|c|c|c|c|}
    			\hline
    			& \textit{Comparison} & \textit{CD} ($\downarrow$) & \textit{E2G} ($\downarrow$) & \textit{G2E} ($\downarrow$) \\
    			\hline
    			{\parbox[t]{2mm}{\multirow{5}{*}{\rotatebox[origin=c]{90}{\textit{Cactus}}}}} & w/o $L_{\text{FLO}}$  & 8.32$\,^\dagger$ & 4.67$\,^\dagger$ & 3.65$\,^\dagger$ \\
    			& w/o $L_{\text{EIK}}$  & 5.47 & 3.58 & 1.89 \\
    			& \multirow{2}{*}{\shortstack{w/o $L_{\text{FLO}}$, $L_{\text{NBR}}$,\\ \& $L_{\text{DIV}}$}} & \multirow{2}{*}{5.34$\,^\dagger$} & \multirow{2}{*}{3.27$\,^\dagger$} & \multirow{2}{*}{2.07$\,^\dagger$} \\
                & & & & \\
    			& \textbf{Ub4D (Ours)} & \textbf{3.06} & 2.42 & 0.64 \\
    			\hline
    			{\parbox[t]{2mm}{\multirow{4}{*}{\rotatebox[origin=c]{90}{\textit{RootTrans}}}}} & w/o $L_{\text{FLO}}$  & 9.42 & 4.83 & 4.59 \\
    			& w/o $L_{\text{EIK}}$  & 0.29 & 0.16 & 0.13 \\
    			& \multirow{2}{*}{\shortstack{w/o $L_{\text{FLO}}$, $L_{\text{NBR}}$,\\ \& $L_{\text{DIV}}$}} & \multirow{2}{*}{4.30$\,^\dagger$} & \multirow{2}{*}{3.17$\,^\dagger$} & \multirow{2}{*}{1.13$\,^\dagger$} \\
                & & & & \\
    			& \textbf{Ub4D (Ours)} & \textbf{0.23} & 0.14 & 0.09 \\
    			\hline
    		\end{tabular}
    	\end{center}

    \label{tab:megatable}
    \vspace{-0.5cm}
	\caption
    {	\label{tab:ablation}
		\textbf{Quantitative ablation study.} 
		``$^\dagger$'' denotes frames 
		that do not produce any geometry (due to frustum culling).
		Note that our full method provides the best result for both scenes.
    }
    \vspace{-0.4cm}
\end{table}

\begin{figure*}[t!]
	\begin{center}
		\includegraphics[width=\textwidth]{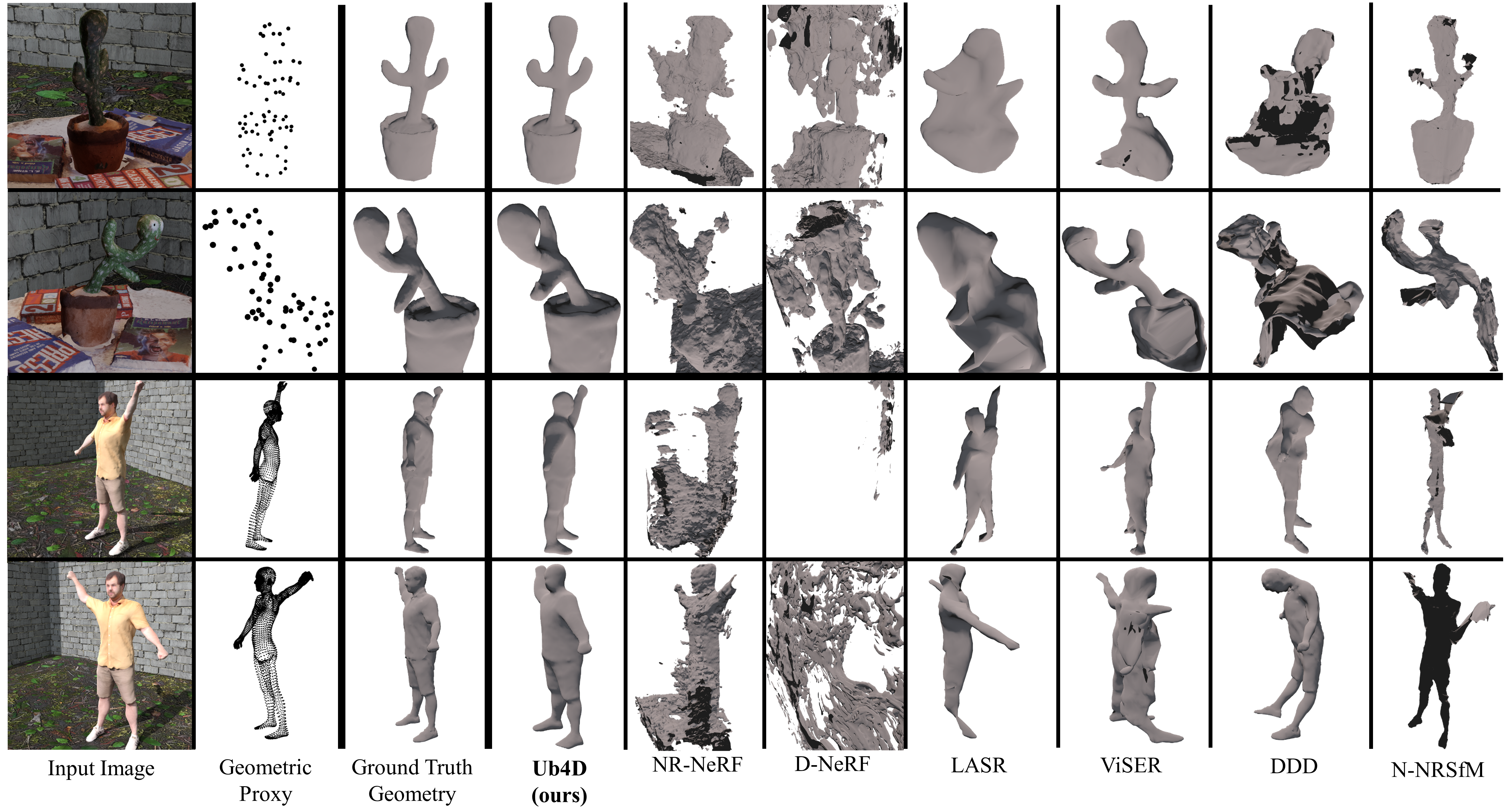}
	\end{center}
    \vspace{-0.5cm}
	\caption
	{
	    Qualitative comparison %
	    with our synthetic sequences rendered from novel views. 
	    Note that competing methods struggle with reconstructing the dense and deforming surface, while Ub4D captures the large scale deformations as well as medium scale details.
	}
	\label{fig:qual_comparison}
    \vspace{-0.5cm}
\end{figure*}
%
%
\subsection{Quantitative Comparison}  \label{sec:quan_compare} 
We compare our method to NR-NeRF~\cite{Tretschk2020}, D-NeRF~\cite{pumarola2020dnerf}, N-NRSfM~\cite{sidhu2020neural}, DDD~\cite{yu2015direct}, LASR~\cite{yang2021lasr}, and ViSER~\cite{yang2021viser}.
NR-NeRF and D-NeRF are novel view synthesis methods that permit extracting geometry by using marching cubes on their density networks with a threshold.
N-NRSfM is \ifglsused{nrsfm}{an}{a} \gls{nrsfm} method, which uses an auto-decoder to deform a mean shape based on a learned per-frame latent representation.
DDD is template-based; it deforms the template to minimize an energy formulation.
For DDD, we provide the first frame's ground-truth mesh as a template. 
Both LASR and ViSER do not require a template and recover a rigged mesh that is animated over the image sequence.
Several other 4D reconstruction techniques with source code available online, such as Shimada \textit{et al.}~\cite{shimada2019ismo} or Ngo \textit{et al.}~\cite{Ngo2015}, do not work under our assumptions; so, we do not include them.
For each related method, we follow the original papers to 
set 
the hyper-parameters.
%
%
\par
The quantitative results on the synthetic sequences are reported in Table~\ref{tab:quan_comparison} and a qualitative comparison is shown in Fig.~\ref{fig:qual_comparison}. 
Ub4D outperforms the state-of-the-arts both quantitatively and qualitatively. 
We found that prior work struggles with large scale deformations resulting in overly noisy results~\cite{pumarola2020dnerf},~\cite{Tretschk2020}, experiences tracking errors~\cite{yu2015direct}, has a limited resolution~\cite{yang2021lasr},~\cite{yang2021viser}, or only reconstructs the visible geometry~\cite{sidhu2020neural} while ours accurately captures the large deformations of the entire geometry.
Also note that although we rigidly align the results for other methods with the ground truth, our method still achieves the most accurate results.
For completeness, we also report our results after ICP.
%
%
\subsection{Comparison to Volume-based Representations} \label{sec:volume_rep}
Like some previous works~\cite{wang2021neus, yariv2020multiview}, our method leverages an SDF representation to model the surface of the object.
An alternative approach is predicting volume density with a network~\cite{Tretschk2020, mildenhall2020nerf, pumarola2020dnerf}.
While a volume density representation has been used for the problem of novel view synthesis, extracting a surface from such a density representation with marching cubes results in noisy and inaccurate geometry as demonstrated in Fig.~\ref{fig:qual_comparison} and further in Fig.~\ref{fig:vs_nrnerf_ablation}-(a) where we qualitatively compare to NR-NeRF~\cite{Tretschk2020}.
In contrast, an SDF representation removes the need to determine a threshold when extracting the explicit geometry and must add a zero crossing, \textit{i.e.,} a surface, in order to satisfy the reconstruction losses.
Further, this example shows the limited ability of NR-NeRF to model large deformations as they penalize the absolute offset length, which our neighbouring frame regularization allows us to handle.
\begin{figure}
        \centering
        \includegraphics[width=\linewidth]{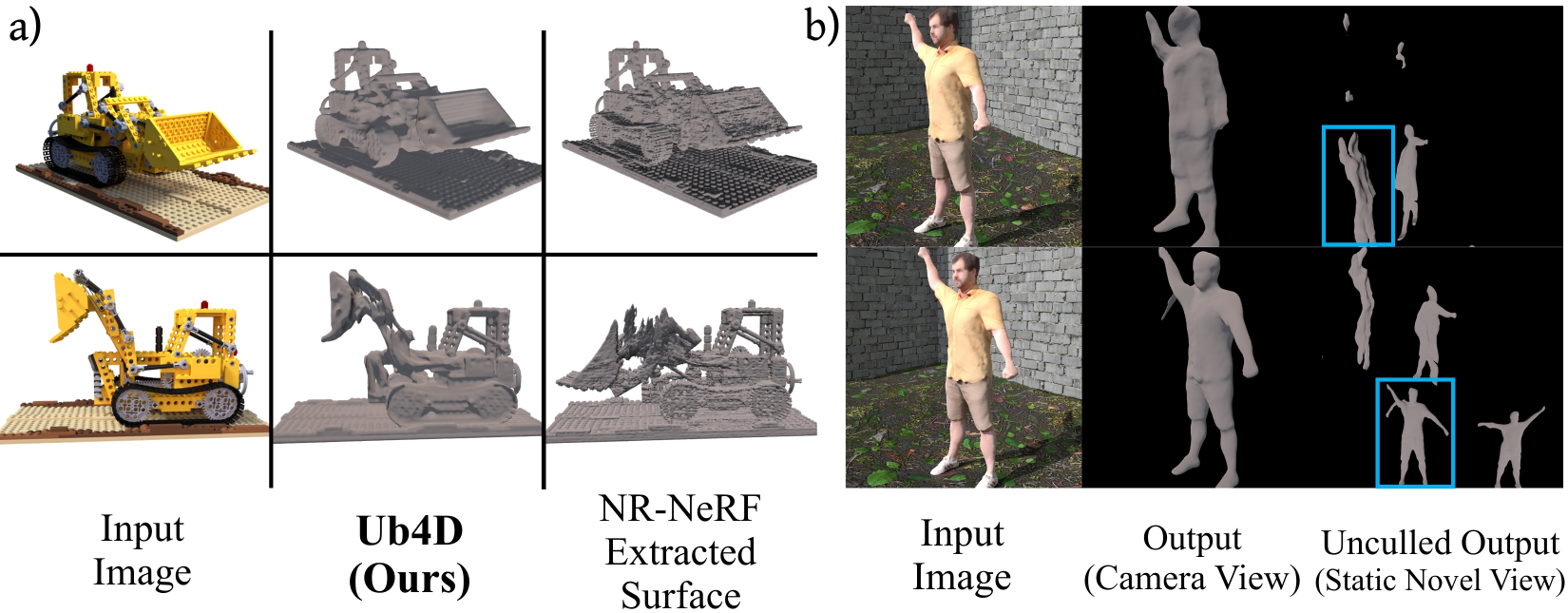}
        \caption{(a) Comparison of Ub4D using an SDF scene representation (without scene flow loss) to a density-based scene representation, called NR-NeRF \cite{Tretschk2020}.
	    To generate surfaces for NR-NeRF, we apply marching cubes~\cite{lorensen1987marching} with a threshold of 50.
	    The density-based representation leads to an overall noisier surface compared to our approach.
	    We also penalize bending using neighbouring frame offsets allowing Ub4D to accurately reconstruct large deformations.
        (b) Qualitative ablation of the scene flow loss ($L_{\text{FLO}}$) on the \textit{RootTrans} sequence.
		Right column shows the scene from a static novel view with the geometry in the camera frustum highlighted with a blue box.
		Note that without the proposed scene flow loss using proxy geometry, Ub4D can produce multiple distinct copies of the character at different scales by exploiting the monocular depth ambiguity.
		}
        \label{fig:vs_nrnerf_ablation}
        \vspace{-0.5cm}
    \end{figure}
%
%

\begin{figure*}
	\begin{center}
		\includegraphics[width=0.90\textwidth]{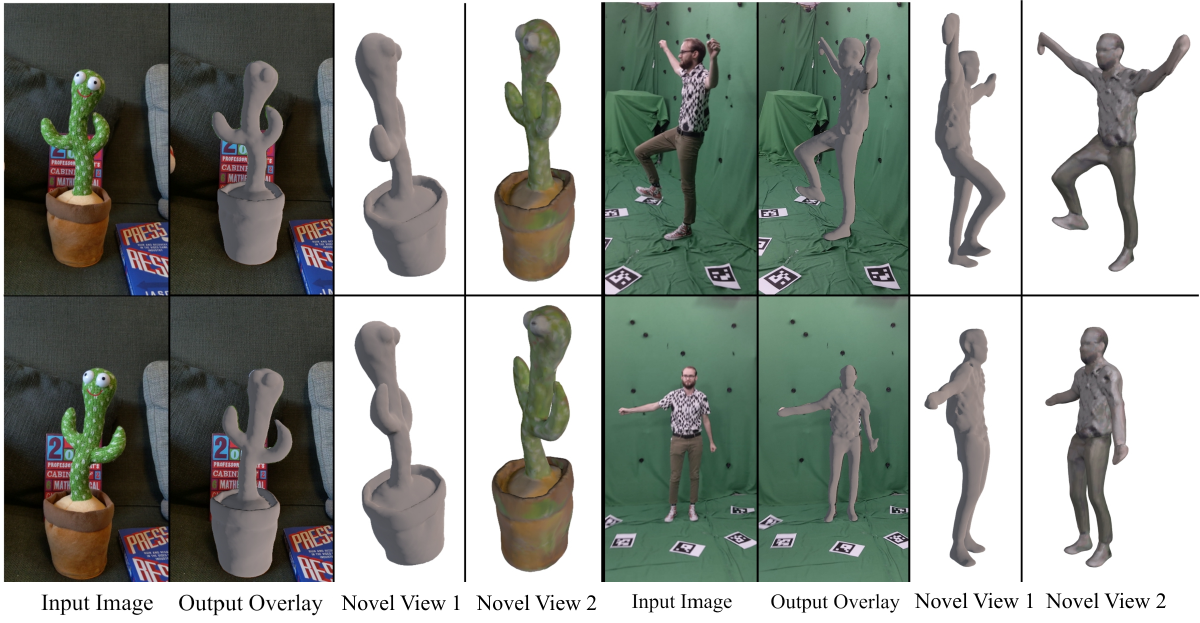}
	\end{center}
    \vspace{-0.59cm}
	\caption
	{
		Qualitative results of our method for a non-humanoid object, \textit{RealCactus}, left, and a human character, \textit{Humanoid}, right.
		The scene flow loss is only used for the human character and only uses a sparse skeleton geometric proxy (12 vertices).
		This shows that the geometric proxy need not be dense or provide any information about the surface. 
		Note that the recovered geometry nicely overlays onto the input image but also looks plausible from a novel 3D viewpoint.
	}
	\label{fig:mega_qualitative}
    \vspace{-0.5cm}
\end{figure*}

\subsection{Ablation Study} \label{sec:ablations}
We validate our design decisions through an ablation study on the \textit{Cactus} and \textit{RootTrans} sequences and report
the metrics 
in Table~\ref{tab:ablation}.  
Our full supervision consists of six loss terms: $L_{\text{COL}}$, $L_{\text{SEG}}$, $L_{\text{EIK}}$, $L_{\text{FLO}}$, $L_{\text{NBR}}$ and $L_{\text{DIV}}$.
We compare the full method to removing the terms: 1) $L_{\text{FLO}}$, which is our novel flow loss,
2) $L_{\text{EIK}}$, which directly regularizes the SDF and color network and indirectly regularizes the bending network  and
3) $L_{\text{FLO}}$, $L_{\text{NBR}}$, and $L_{\text{DIV}}$, which are all direct bending network regularizers.
Most importantly, the full combination of losses provides the best result validating the contribution of each term. 
\par 
Concerning 1), our flow loss especially helps for the large root translation and arm motion of the \textit{RootTrans} sequence.
Without using this loss, multiple different geometries are synthesized, which fit the reconstruction losses.
Then, the bending network can ``switch" between the different copies throughout the sequence.
This results in overfitting to the camera pose by exploiting monocular depth ambiguities to generate geometry that is not seen in other views.
Figure~\ref{fig:vs_nrnerf_ablation}-(b) shows this overfitting to the camera pose with multiple distinct geometries being used over the sequence to satisfy the reconstruction losses.
\par 
Regarding 2), we found that not using $L_{\text{EIK}}$ leads to overall noisier surfaces and thus the quality is reduced.
Finally concerning 3), without any explicit regularization of the bending network, the deformations can be almost arbitrary again leading to overfitting to individual frames by violating 3D consistency resulting in a reduced accuracy.
We even observed that the network was not able to produce any geometry for some frames, which validates the necessity of explicit regularization of the bending network.
%
%
\subsection{Qualitative Results on Real Word Scenes}\label{sec:qual_results} 
We next demonstrate that Ub4D works well on real-world scenes. 
Fig.~\ref{fig:mega_qualitative} visualizes our \textit{RealCactus} sequence depicting a dancing cactus toy 
and the \textit{Humanoid} sequence where a person is moving their arms and legs.
Although in both cases the dynamic scenes contain large deformations, Ub4D robustly and accurately reconstructs individual frame geometries, which also contain medium frequency details. 
%
%
%
\section{Discussion and Possible  Extensions}\label{sec:limitations} 

Ub4D significantly outperforms all competing methods in our evaluations, both numerically and qualitatively. 
We hypothesise that it is partially due to its shape completion property.
In fact, we found that none of the existing methods can deal with captures that include object motion \textit{and} severe camera motions while our method leverages such recording conditions to its benefit, inspired by classical non-rigid structure from motion algorithms.
In the case of severe scene deformations Ub4D can leverage a geometric proxy.%
Note that our scene flow loss is versatile: The proxy can be either a full estimated mesh or even just a few points (see \textit{Humanoid} results Fig.~\ref{fig:mega_qualitative}); as long as it \textit{roughly} describes the deformation of the scene, the model is guided towards a better local minimum in the reconstruction losses.
Future work involves exploring this direction further with the main question being: How sparse can the proxy be and could it even be a 2D entity in the image plane? 
Along these lines, we see multiple avenues for future research, including tracking a generic proxy along with learning the SDF and using 2D image features for initializing a sparse proxy.
%
 
\section{Conclusion}\label{sec:conclusion}  
We presented, Ub4D, a method of a new class for 3D reconstruction of deformable scenes from a single RGB camera. 
It represents the scene as a learned static canonical volume with an implicit surface. 
A bending network warping the frames into this canonical volume accounts for the scene deformation.
Our optional scene flow loss improves the  reconstruction  accuracy and robustness in the case of large deformations given a coarse proxy geometry. 
The qualitative and quantitative comparisons to different method types show that our approach is a clear step towards dense and deformable tracking of general and largely deforming scenes using a single RGB camera. 
\section*{Acknowledgement}
This work was funded by the ERC Consolidator Grant 4DRepLy (770784).
{\small 
\bibliographystyle{ieee_fullname} 
\bibliography{egbib.bib} 
} 
 
\makeatletter
\twocolumn
\appendix 
\noindent{\fontsize{15}{15}\selectfont \textbf{Appendices}}
\vspace{0.3cm}

In this supplementary material, we present additional results in Section~\ref{sec:additional_results} and implementation details in Section~\ref{sec:impl_details}.
Section~\ref{sec:derivation} provides a derivation of the discrete opacity equation for bent rays and Section~\ref{sec:proof} shows the unbiased nature of our rendering method.
We provide a description of the metrics used in our quantitative analysis in Section~\ref{sec:quantitative_metrics}.
Additional details on the experiments performed are given in Section~\ref{sec:experimental_details}, including how we compute the geometric proxies for our human character scenes from the input images in Section~\ref{sec:hgp}.
Section~\ref{sec:proxy_ablation} investigates the necessary accuracy and resolution of the geometric proxies.
Section~\ref{sec:latents} analyzes the learned latent codes and demonstrates novel geometry synthesis.
Finally, we discuss some additional limitations in Section~\ref{sec:additional_limitations}.
%
%
\section{Additional Results} \label{sec:additional_results}
Figures \ref{fig:supp_cactus} and \ref{fig:supp_roottrans} show additional results for our synthetic sequences: \textit{i.e.,} \textit{Cactus} and \textit{RootTrans}, respectively.
We include a qualitative visualization of the output's Chamfer distance to the ground truth where dark blue is zero and red is above the mid-point between the mean and the maximum, both taken over the entire sequence.
Note that overall our reconstructed surface has a very low error and only a small region was not precisely reconstructed.
Similar to many monocular reconstruction methods, the majority of our higher error regions are due to the monocular depth ambiguity (rows 1 and 3 in Figure \ref{fig:supp_cactus}; and row 4 in Figure \ref{fig:supp_roottrans}).
For further results, we also direct the reader to the video.
\suppfig{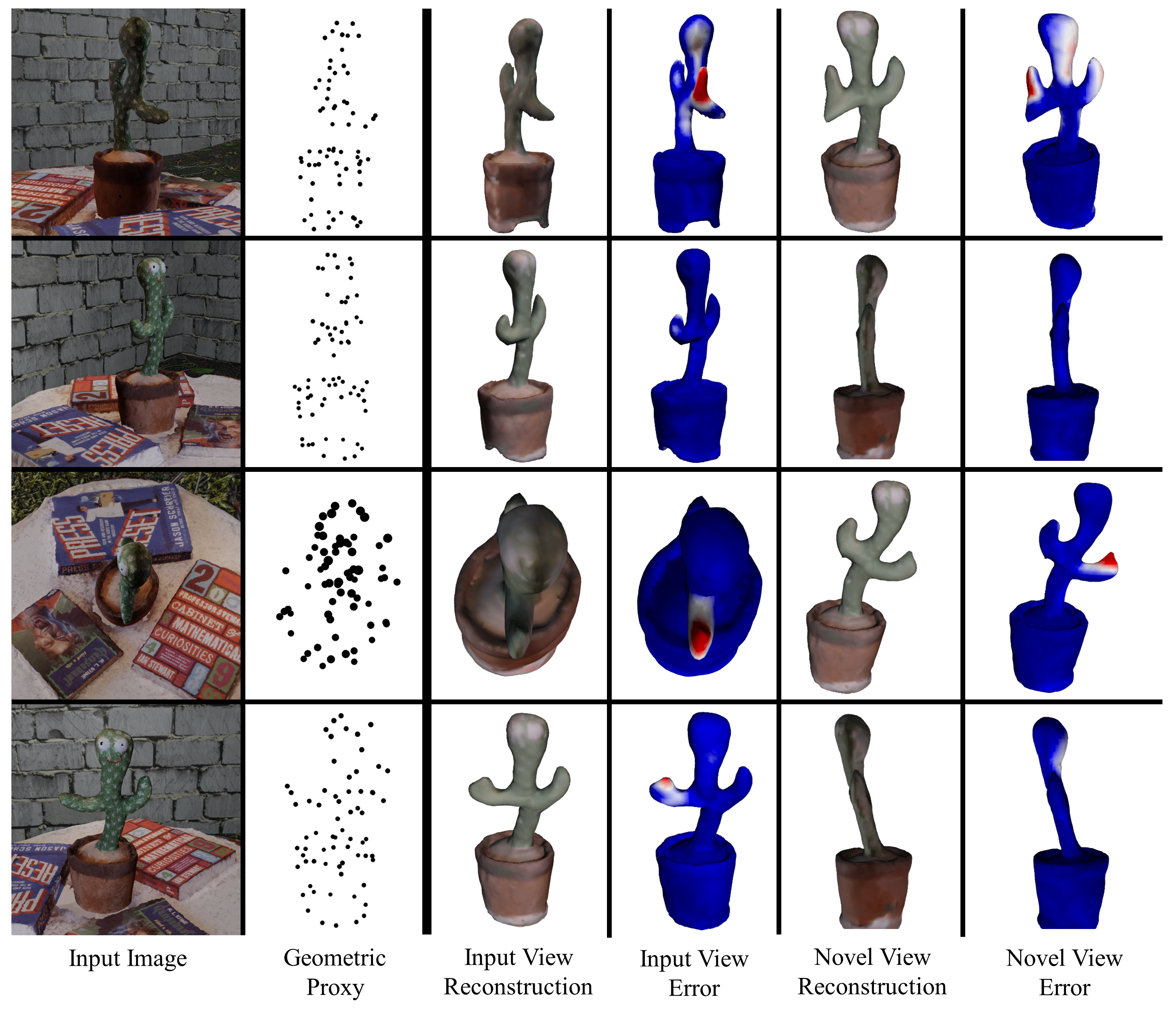}{Additional results for our \textit{Cactus} sequence. We include error coloring where blue is low error and red is high error, relative to the entire sequence.}{fig:supp_cactus}
\suppfig{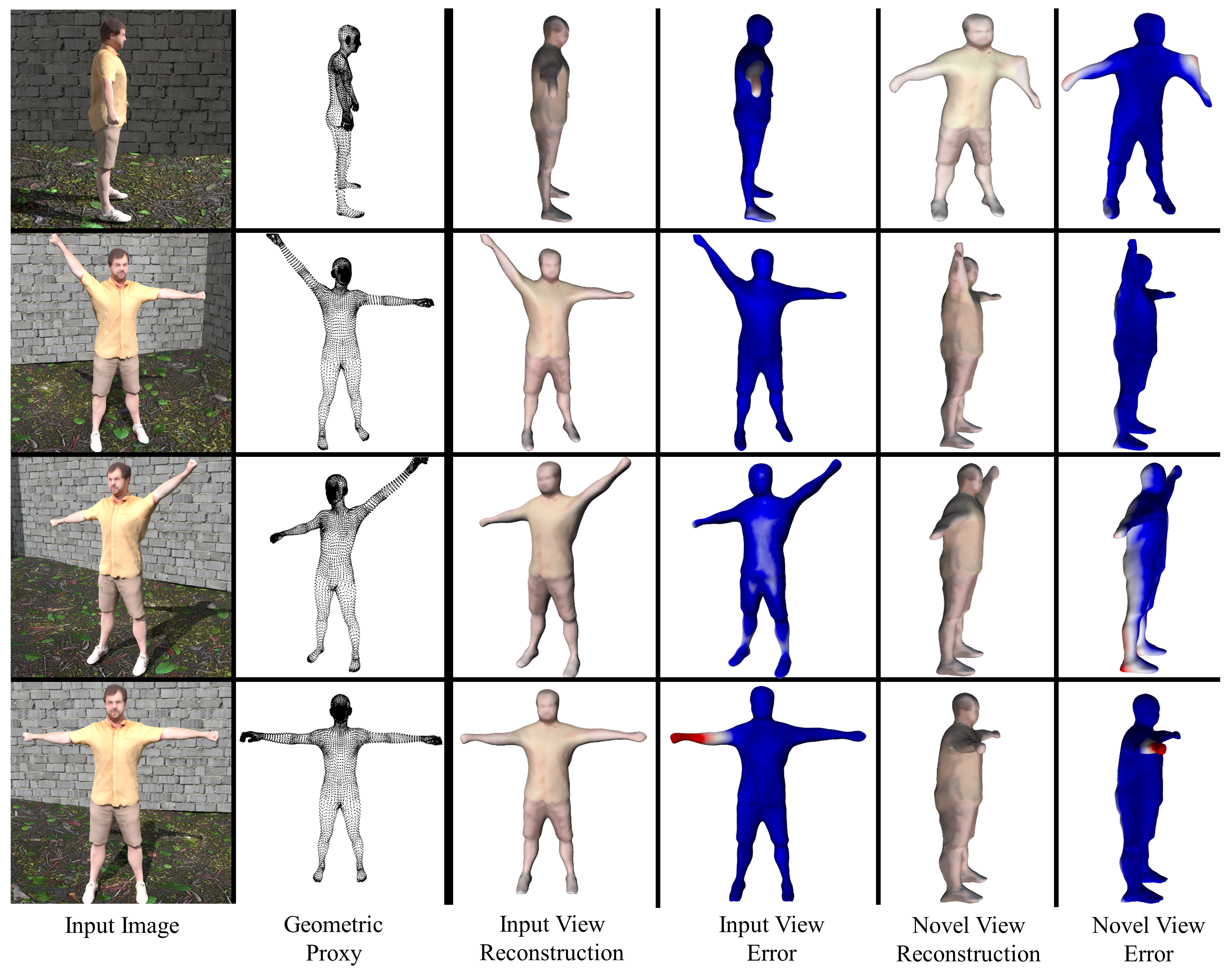}{Additional results for our \textit{RootTrans} sequence. We include error coloring where blue is low error and red is high error, relative to the entire sequence.}{fig:supp_roottrans}
%
%
\section{Implementation Details} \label{sec:impl_details}
We base our implementation on the codebase of Wang~et~al.~\cite{wang2021neus}, which is implemented in PyTorch~\cite{pytorch}.
Our method consists of 3 \gls{mlp} networks: \textit{Bending}, \textit{SDF}, and \textit{Rendering}.
We provide a diagram showing the networks in Figure \ref{fig:network_diagram} and give additional details in Table \ref{tab:network_parameters}.
We also configure the starting weights of \textit{SDF} using the geometric initialization method of Atzmon and Lipman~\cite{atzmon2020sal}.
\par
At the start of training we follow Tretschk~et~al.~\cite{Tretschk2020} and initialize the latent codes with zeros.
For each iteration during training, we select $512$ pixels uniformly over the image for which to fire rays.
We sample $64$ positions along each straight ray, jitter these samples, and then importance sample $64$ additional positions based on the \gls{sdf} values.
\begin{figure}[ht!]
	\begin{center}
		\includegraphics[width=0.9\linewidth]{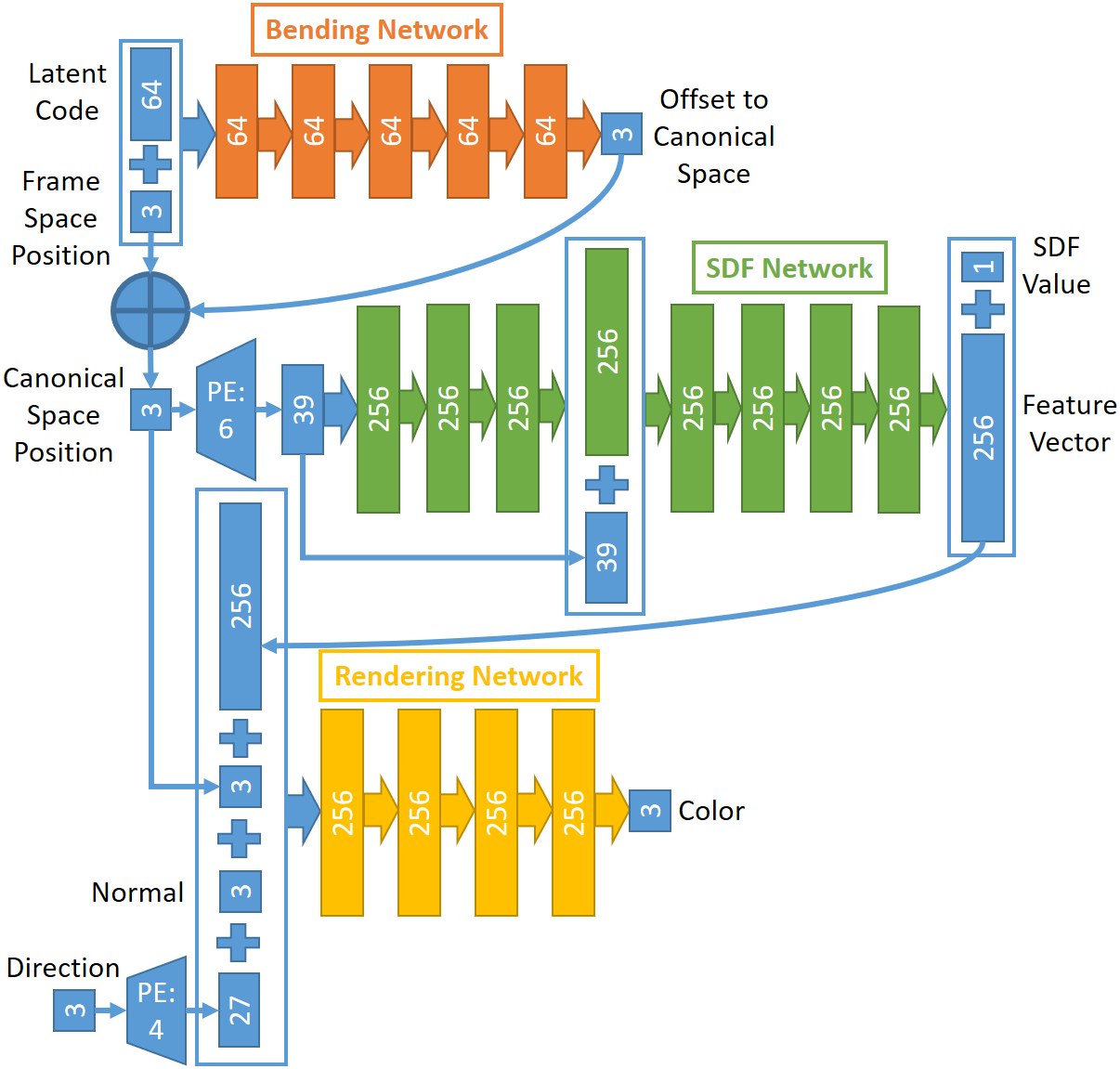}
	\end{center}
	\caption
	{
	    Network diagram of our method.
	    PE denotes Positional Encoding~\cite{mildenhall2020nerf} with the given number of additional frequencies.
	    The addition node performs element-wise addition while the plus blocks represent vector concatenation.
	}
	\label{fig:network_diagram}
\end{figure}
{\renewcommand{\arraystretch}{1.1}
\begin{table}[!ht]
	\begin{center}
		\begin{tabular}{|c|c|c|}
			\hline
			\textit{Network} & \textit{Activation} & \textit{Weight Normalization} \\
			\hline
			\textit{Bending} & ReLU & No \\
			\textit{SDF} & SoftPlus ($\beta = 100$) & Yes \\
			\textit{Rendering}  & ReLU & Yes \\
			\hline
		\end{tabular}
	\end{center}
	\caption
	{
	    Network parameters used in our implementation.
	}
	\label{tab:network_parameters}
\end{table}
}
%
%
\section{Derivation of the Discrete Opacity $\alpha_i^{(z)}$ Equation for Bent Rays} \label{sec:derivation}
In this section we show that the derivation of discrete opacity $\alpha_i^{(z)}$ (Section~3.2) follows from volume rendering principles and the definition of the opaque density $\rho_i (\Tilde{\vv{r}}(t))$ (Section~3.2, Equation~3).
This analysis is similar to that in the Appendix~A of Wang~et~al.~\cite{wang2021neus}, where we extend their derivation to any smooth parametric path.
\par
Before beginning, we remind the reader that we render along bent rays, which are parametric paths in $\mathbb{R}^3$:
\begin{align}
    \br{i}{t} &= \vv{r}(t) + \vv{b}_i\big(\vv{r}(t)\big),\,\,\, \vv{r}(t) = \vv{o} + t\vv{d}
\end{align}
We direct the reader to Section~3.1 for the definition of these terms.
At each point on the bent ray there is an instantaneous viewing direction $\od{\br{i}{t}}{t}$ which can be computed analytically as:
\begin{align}
    \od{\br{i}{t}}{t} &= \od{}{t}\big[ \vv{r}(t) \big] + \od{}{t}\big[ \vv{b}_i(\vv{r}(t)) \big] \\
    &= \vv{d} + \frac{\partial \vv{b}_i}{\partial \vv{r}(t)} \od{}{t}\big[ \vv{r}(t) \big] \\
    &= \vv{d} + \frac{\partial \vv{b}_i}{\partial \vv{r}(t)} \vv{d}
\end{align}
where $\frac{\partial \vv{b}_i}{\partial \vv{r}(t)}$ is the Jacobian of the bending network w.r.t. its input $\vv{r}(t)$, a point along the straight ray.
Note that, in our case, the Jacobian exists everywhere since the bending network is \ifglsused{mlp}{an}{a} \gls{mlp}; thus our bent ray is a smooth parametric path.
\par
In Section~3.2, we define the opaque density as:
\begin{align} \label{eqn:opaque_density_repeat}
    \rho_i(\Tilde{\vv{r}}_i (t)) &= \text{max} \Bigg\{ \frac{-\od{\Phi_s}{t}\big( f(\Tilde{\vv{r}}_i(t)) \big)}{\Phi_s \big( f(\Tilde{\vv{r}}_i(t)) \big)}, 0 \Bigg\}
\end{align}
where $\Phi_s$ is the \gls{cdf} of the logistic distribution.
In order to proceed, we must expand the numerator through the chain rule:
\begin{align}
    \od{\Phi_s}{t}\big( f(\br{i}{t}) \big) &= \phi_s \big( f(\br{i}{t}) \big) \od{}{t} \big( f(\br{i}{t} \big) \\
    &= \phi_s \big( f(\br{i}{t}) \big) \big[\nabla f(\br{i}{t}) \cdot \od{\br{i}{t}}{t} \big] \label{eqn:expanded_numerator}
\end{align}
where $\phi_s$ is the \gls{pdf} of the logistic distribution.
There is no need to expand the instantaneous viewing direction now that we have demonstrated its smoothness.
\par
Placing Equation~\ref{eqn:expanded_numerator} into Equation~\ref{eqn:opaque_density_repeat}, gives:
\begin{align}
    &\rho_i(\Tilde{\vv{r}}_i (t))= \\ 
    & \text{max} \Bigg\{ -\frac{\phi_s \big( f(\br{i}{t}) \big) \big[\nabla f(\br{i}{t}) \cdot \od{\br{i}{t}}{t} \big]}{\Phi_s \big( f(\Tilde{\vv{r}}_i(t)) \big)}, 0 \Bigg\}
\end{align}
There are two regions of interest identified in Appendix~A of NeuS~\cite{wang2021neus}: a ray entering geometry and a ray exiting geometry.
\par
\begin{figure}[t!]
	\begin{center}
		\includegraphics[width=0.3\linewidth]{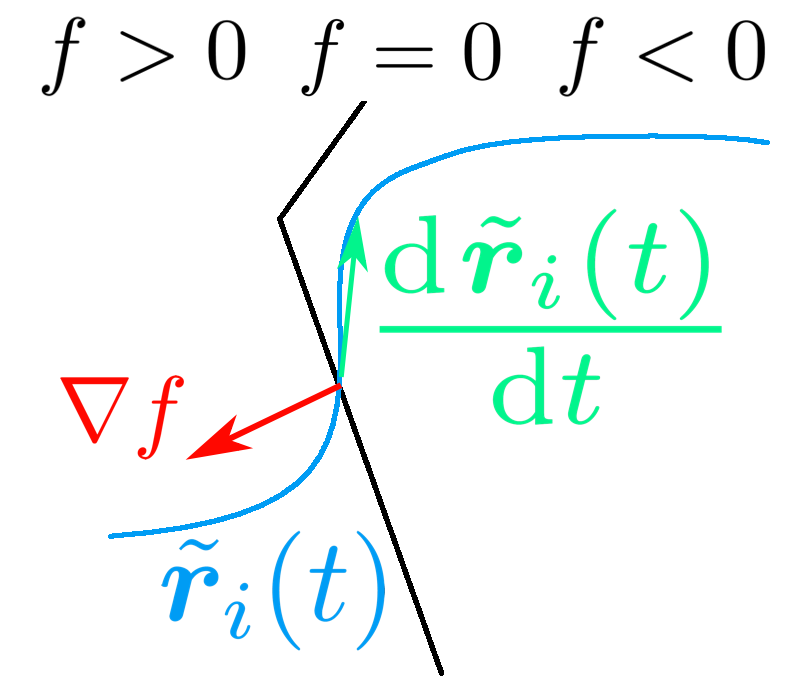}
	\end{center}
	\caption
	{
	    Graphical depiction of a bent ray (traveling left to right) entering \ifglsused{sdf}{an}{a} \gls{sdf} surface. Note that the instantaneous viewing direction and gradient of the \gls{sdf} must have a negative dot product for the bent ray to be entering the geometry.
	}
	\label{fig:neus_entering_surface}
\end{figure}
We first present the case where a ray is entering the geometry as depicted in Figure~\ref{fig:neus_entering_surface}.
Since we know that in this case:
\begin{align}
    \nabla f(\br{i}{t}) \cdot \od{\br{i}{t}}{t}  < 0
\end{align}
and that both $\phi_s$ and $\Phi_s$ are non-negative, we can drop the maximum in the opaque density and return the numerator to its more condensed form:
\begin{align}
    \rho_i(\Tilde{\vv{r}}_i (t)) &= \frac{-\phi_s \big( f(\br{i}{t}) \big) \Big[\nabla f(\br{i}{t}) \cdot \od{\br{i}{t}}{t} \Big]}{\Phi_s \big( f(\Tilde{\vv{r}}_i(t)) \big)} \\
    &= \frac{-\od{\Phi_s}{t}\big( f(\Tilde{\vv{r}}_i(t)) \big)}{\Phi_s \big( f(\Tilde{\vv{r}}_i(t)) \big)}
\end{align}
The remaining derivation for the discrete opacity $\alpha_i^{(z)}$ follows exactly as in Appendix A of NeuS~\cite{wang2021neus}:
\begin{align*}
    &\alpha_i^{(z)} = 1-\exp\left(-\int_{t^{(z)}}^{t^{(z+1)}}\rho(t){\rm d}t\right) \\
    &= 1-\exp\left(-\int_{t^{(z)}}^{t^{(z+1)}}\frac{-\od{\Phi_s}{t}\big( f(\Tilde{\vv{r}}_i(t)) \big)}{\Phi_s \big( f(\Tilde{\vv{r}}_i(t)) \big)}{\rm d}t\right) \\
    &= 1-\exp\left( \ln \big[ \Phi_s \big( f(\br{i}{t^{(z+1)}} \big) \big] - \ln \big[ \Phi_s \big( f(\br{i}{t^{(z)}} \big) \big] \right) \\
    &= 1-\frac{\Phi_s \big( f(\br{i}{t^{(z+1)}} \big)}{\Phi_s\big( f(\br{i}{t^{(z)}} \big)} \\
    &= \frac{\Phi_s\big( f(\br{i}{t^{(z)}} \big) - \Phi_s\big( f(\br{i}{t^{(z+1)}} \big)}{ \Phi_s\big( f(\br{i}{t^{(z)}} \big)} \label{eqn:alpha_i_last_line}
\end{align*}
Note that Equation~\ref{eqn:alpha_i_last_line} is non-negative ($\because f(\br{i}{t^{(z)}} > f(\br{i}{t^{(z+1)}}$ and $\Phi_s$ is non-negative and monotonically increasing) and as such is equivalent to a maximum with zero.
\par
The second case to consider is where a ray is exiting the geometry as depicted in Figure~\ref{fig:neus_exiting_surface}.
Given that:
\begin{align}
    \nabla f(\br{i}{t}) \cdot \od{\br{i}{t}}{t}  > 0
\end{align}
and that both $\phi_s$ and $\Phi_s$ are non-negative, Equation \ref{eqn:opaque_density_repeat} gives that $\rho_i (\br{i}{t}) = 0$.
Thus the discrete opacity $\alpha_i^{(z)}$ is:
\begin{align}
    \alpha_i^{(z)} &= 1-\exp\left(-\int_{t^{(z)}}^{t^{(z+1)}}\rho(t)\,{\rm d}t\right) \\
    &= 1-\exp\left(-\int_{t^{(z)}}^{t^{(z+1)}}0\,{\rm d}t\right) \\
    &= 0
\end{align}
Since Equation~\ref{eqn:alpha_i_last_line} will be non-positive when exiting the geometry ($\because f(\br{i}{t^{(z+1)}} > f(\br{i}{t^{(z)}}$ and $\Phi_s$ is non-negative and monotonically increasing), we can write the derived equation for $\alpha_i^{(z)}$ satisfying both cases as:
\begin{align}
    &\alpha_i^{(z)} = \\
    &\text{max} \Bigg\{ \frac{\Phi_s\Big(f\big(\Tilde{\vv{r}}_i(t^{(z)})\big)\Big) - \Phi_s\Big(f\big(\Tilde{\vv{r}}_i(t^{(z+1)})\big)\Big)}{\Phi_s\Big(f\big(\Tilde{\vv{r}}_i(t^{(z)})\big)\Big)}, 0 \Bigg\}
\end{align}
\begin{figure}[t!]
	\begin{center}
		\includegraphics[width=0.3\linewidth]{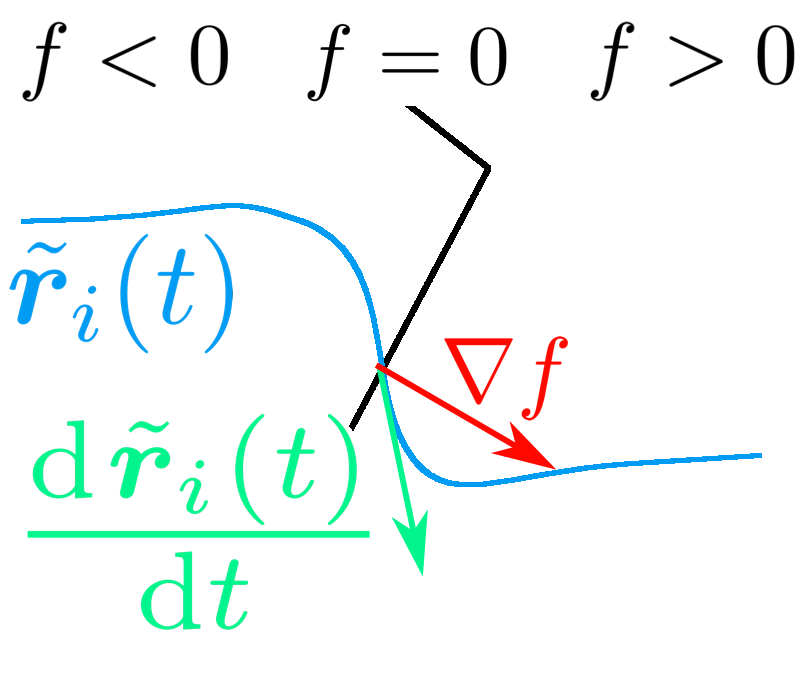}
	\end{center}
	\caption
	{
	    Graphical depiction of a bent ray (traveling left to right) exiting \ifglsused{sdf}{an}{a} \gls{sdf} surface. Note that the instantaneous viewing direction and gradient of the \gls{sdf} must have a positive dot product for the bent ray to be exiting the geometry.
	}
	\label{fig:neus_exiting_surface}
\end{figure}
%
%
\section{Unbiased Nature of our Rendering Method} \label{sec:proof}
In this section, we show that our rendering method is unbiased with respect to the surface of the object, i.e. $f\big(\bri{t}\big) = 0$, given that $s$ becomes sufficiently small.
This demonstration follows a similar progression to that in the Appendix~B of Wang~et~al.~\cite{wang2021neus}.
We assume two theoretical properties: that $f$ is an SDF and that $\od{\bri{t}}{t}$ is never the zero vector.
Both of these properties are enforced by penalizers in our method (i.e. the Eikonal and divergence regularizers respectively), but are not strictly guaranteed.
Specifically, the $\od{\bri{t}}{t} \neq \vv{0}$ property follows from a divergence-free bending network since this prevents the compression of space necessary to give a stationary ray.
\par
From Figure~\ref{fig:neus_entering_surface}, it can be seen that for a smooth parametric path to intersect the surface, there \textit{must} be a finite region $t \in (t_l, t_r)$ such that $\nabla f \big(\bri{t}\big) \cdot \od{\bri{t}}{t} < 0$.
We can re-write the weight as:
\footnotesize
\begin{align*}
    &\omega(\ri, t) = T(\ri, t)\, \rho(\bri{t}) \\
    &= \exp\left(-\int_{0}^{t} \rho(\bri{\tau})\,{\rm d}\tau\right) \rho(\bri{t}) \\
    &= \exp\left(-\int_{0}^{t_l} \rho(\bri{\tau})\,{\rm d}\tau\right) \exp\left(-\int_{t_l}^{t} \rho(\bri{\tau})\,{\rm d}\tau\right) \rho(\bri{t}) \\
    &= T(\ri, t_l) \exp\left( \ln\big[ \Phi_s\big(f(\bri{t})\big) \big] - \ln\big[ \Phi_s\big( f(\bri{t_l} \big) \big]\right) \rho(\bri{t}) \\
    &= T(\ri, t_l) \frac{\cancel{\Phi_s\big(f(\bri{t})\big)}}{\Phi_s\big(f(\bri{t_l})\big)} \frac{\Big[-\nabla f(\bri{t}) \cdot \od{\bri{t}}{t} \Big]\, \phi_s \big( f(\bri{t}) \big)}{\cancel{\Phi_s \big( f(\bri{t}) \big)}} \\
    &\therefore \omega(\ri, t) = \underbrace{\frac{T(\ri, t_l)}{\Phi_s\big(f(\bri{t_l})\big)}}_{\text{constant w.r.t. }t} \underbrace{\Big[-\nabla f(\bri{t}) \cdot \od{\bri{t}}{t} \Big]\, \phi_s \big( \f(\bri{t}) \big)}_{F(t)}\, .
\end{align*}
\normalsize
Then we can establish that for:
\begin{align}
    F(t) &= \underbrace{\Big[-\nabla f(\bri{t}) \cdot \od{\bri{t}}{t} \Big]}_{G(t)}\, \phi_s \big( f(\bri{t}) \big)\, ,
\end{align}
$\exists s>0$ such that $F(t)$ is maximized by $\f(\bri{t^*}) = 0$, $t^* \in (t_l, t_r)$.
Consider another value $t^\dagger \in (t_l, t_r)$, $t^\dagger \neq t^*$ where $G(t^\dagger) = 1$ is maximum and $G(t^*) = \epsilon$ is minimum for some necessarily non-zero value $\epsilon$.
This corresponds to the worst case for the unbiasedness since $0 < G(t) \leq 1$, $\forall t \in (t_l, t_r)$.
Then:
\begin{align}
    G(t^*)\, \phi_s \big( \f(\bri{t^*}) \big) \stackrel{?}{>}&\,\, G(t^\dagger)\, \phi_s \big( \f(\bri{t^\dagger}) \big) \\
    \frac{\phi_s (0)}{\phi_s \big( \f(\bri{t^\dagger}) \big)} \stackrel{?}{>}&\,\, \frac{G(t^\dagger)}{G(t^*)} = \frac{1}{\epsilon}\, . \label{eqn:ratio}
\end{align}
Taking the limit of the left-hand side of Equation~\eqref{eqn:ratio} as $s$ approaches $0$ and using the definition of the logistic PDF $\phi_s$:
\begin{align}
    &\lim_{s \rightarrow 0} \frac{\phi_s (0)}{\phi_s \big( \f(\bri{t^\dagger}) \big)} \\
    &= \lim_{s \rightarrow 0} \frac{\exp\left(\frac{\f(\bri{t^\dagger})}{s}\right)}{4\left(1 + \exp\left(-\frac{\f(\bri{t^\dagger})}{s}\right)\right)^2} \\
    &= \infty\, .
\end{align}
Thus, for every possible $\epsilon$, $\exists s > 0$ such that:
\begin{gather}
    \frac{\phi_s (0)}{\phi_s \big( \f(\bri{t^\dagger}) \big)} > \frac{1}{\epsilon}\, ,
\end{gather}
which implies $F(t^*) > F(t^\dagger),\, \forall t^\dagger \in (t_l, t_r),\, t^\dagger \neq t^*$. \hfill$\square$
%
%
\section{Quantitative Metrics} \label{sec:quantitative_metrics}
This section presents the quantitative metrics used in our paper in detail.
We use the Chamfer Distance (CD) as defined below:
\begin{align} \label{eqn:cd_full}
\begin{split}
    \text{CD}(E, G) &= \text{E2G}(E, G) + \text{G2E}(E, G) \\
    &= \frac{1}{|E|} \sum_{\vec{x} \in E} \min_{\vec{y} \in G} ||\vec{x} - \vec{y}||_2^2 \\
    &\hspace{0.5cm}+ \frac{1}{|G|} \sum_{\vec{y} \in G} \min_{\vec{x} \in E} ||\vec{y} - \vec{x}||_2^2 \\
\end{split}
\end{align}
where $E$ is the estimated mesh and $G$ is the ground truth mesh.
In addition, we also report the unidirectional parts shown above individually due to the difference in methods to which we compare:
\begin{align} \label{eqn:quan_parts}
    \text{E2G}(E, G) &= \frac{1}{|E|} \sum_{\vec{x} \in E} \min_{\vec{y} \in G} ||\vec{x} - \vec{y}||_2^2, \\
    \text{G2E}(E, G) &= \frac{1}{|G|} \sum_{\vec{y} \in G} \min_{\vec{x} \in E} ||\vec{y} - \vec{x}||_2^2.
\end{align}
\par
As mentioned in the paper, the CD metric is not always fair for each method which is why we also report the E2G and G2E values separately.
The E2G metric is unfair for the novel view synthesis methods, that is NR-NeRF \cite{Tretschk2020} and D-NeRF \cite{pumarola2020dnerf}.
This is because we must select a region in which to employ marching cubes for generating geometry and this selection affects the E2G metric.
The G2E metric is unfair for N-NRSfM \cite{sidhu2020neural} since this method only reconstructs the visible surface of the object.
%
%
\section{Experimental Details} \label{sec:experimental_details}
We present hyperparameters and additional details for the experiments with respect to our method and previous works.
Sections~\ref{sec:ub4d}, \ref{sec:lasr}, \ref{sec:ddd}, and \ref{sec:nnrsfm} give the experimental details for Ub4D, LASR, DDD, and N-NRSfM, respectively. 
As a reminder, the datasets used are summarized in Table~\ref{tab:datasets}.
{\renewcommand{\arraystretch}{1.1}
\begin{table*}[ht!]
	\begin{center}
		\begin{tabular}{|c|c|c|c|c|c|}
			\hline
			\textit{Name} & \textit{Creation} & \textit{Frames} & \textit{Resolution} & \textit{Geometric Proxies?} & \textit{GT?} \\
			\hline
			\textit{Cactus} & Blender~\cite{Hess2010} & $150$ & $1024{\times}1024$ & Yes (decimated GT) & Yes \\
			\textit{RootTrans} & Blender~\cite{Hess2010} & $150$ & $1024{\times}1024$ & Yes (SMPL~\cite{SMPL-X:2019}) & Yes \\
			\textit{Lego} & Blender~\cite{Hess2010} & $150$ & $800{\times}800$ & No & No \\
			\hdashline
			\textit{Humanoid} & Real World & $171$ & $960{\times}1280$ & Yes (SMPL~\cite{SMPL-X:2019}) & No \\
			\textit{RealCactus} & Real World & $150$ & $1080{\times}1920$ & No & No \\
			\hline
		\end{tabular}
	\end{center}
	\caption[Summary of the datasets introduced in this work.]
	{
		Summary of the datasets introduced in this work.
		GT indicates access to ground truth geometry in the form of per-frame meshes.
		Synthetic scenes are above the dashed line, real-world captures below.
	}
	\label{tab:datasets}
\end{table*}
}
\subsection{Ub4D} \label{sec:ub4d}
The hyperparameter settings for the experiments presented are contained in Table~\ref{tab:experimental_parameters}.
Our loss weights are all relative to the colour weight:
\begin{align} \label{eqn:full_loss}
\begin{split}
    L &= L_{\text{COL}} + \omega_{\text{SEG}} L_{\text{SEG}} + \omega_{\text{EIK}} L_{\text{EIK}} + \omega_{\text{NBR}} L_{\text{NBR}} \\
      &+ \omega_{\text{DIV}} L_{\text{DIV}} + \omega_{\text{FLO}} L_{\text{FLO}}
\end{split}
\end{align}
Additionally, for the \textit{RealCactus} experiment we use constant $\omega_{\text{NBR}}$ and $\omega_{\text{DIV}}$ weights, rather than the $\frac{1}{100}$ factor exponentially increasing schedule of Tretschk et al.~\cite{Tretschk2020}.
\par
We give the time to apply marching cubes over the scene, which is independent of scene, in Table~\ref{tab:marching_time}.
This is given \textit{without} applying frustum culling since that is an insignificant portion of the total runtime and is scene and region dependent.
{\renewcommand{\arraystretch}{1.1}
\begin{table*}[!ht]
	\begin{center}
		\begin{tabular}{|c|c|c|c|c|c|c|c|c|c|}
			\hline
			\textit{Scene} & \textit{Iterations} (1000) & \textit{Training} (hours) & $\omega_{\text{SEG}}$ & $\omega_{\text{EIK}}$ & $\omega_{\text{NBR}}$ & $\omega_{\text{DIV}}$ & $\omega_{\text{FLO}}$ & $\lambda_1$ & $\lambda_2$ \\
			\hline
			\textit{Cactus} & 300 & 17.0 & 1.0 & 0.5 & 20000 & 200 & 10 & 700 & 75 \\
			\textit{RootTrans} & 450 & 26.4 & 1.0 & 0.5 & 20000 & 200 & 10 & 700 & 75 \\
			\textit{Lego} & 450 & 19.9 & 0.75 & 0.25 & 10000 & 100 & 0 & - & - \\
			\textit{RealCactus} & 450 & 22.1 & 1.5 & 0.75 & 15000$^\dagger$ & 50$^\dagger$ & 0 & - & - \\
			\textit{Humanoid} & 450 & 21.5 & 1.25 & 0.25 & 50000 & 200 & 10 & 700 & 75 \\
			\hline
		\end{tabular}
	\end{center}
	\caption
	{
	    Hyperparameters used in acquiring results presented. $^\dagger$ indicates a constant weight without the increasing schedule of Tretschk~et~al.~\cite{Tretschk2020}.
	}
	\label{tab:experimental_parameters}
\end{table*}
}
{\renewcommand{\arraystretch}{1.1}
\begin{table}[!ht]
	\begin{center}
		\begin{tabular}{|c|c|c|}
			\hline
			\textit{Resolution}  & \multicolumn{2}{c}{\textit{March Time}} \vline \\
			& (seconds) & (hours) \\
			\hline
			64 & 14 & 0.00 \\
			128 & 69 & 0.02 \\
			256 & 536 & 0.15 \\
			512 & 4224 & 1.17 \\
			1024 & 34.99\e{6} & 18.32 \\
			\hline
		\end{tabular}
	\end{center}
	\caption
	{
	    Time required to march geometry (without frustum culling).
	}
	\label{tab:marching_time}
\end{table}
}
\subsection{NR-NeRF \cite{Tretschk2020}} \label{sec:nrnerf}
We run NR-NeRF~\cite{Tretschk2020} in the manner shown in their code\footnote{\url{https://github.com/facebookresearch/nonrigid_nerf}}.
We use the default parameters provided by the authors and sweep the threshold to find the threshold giving minimal metric (5 for the \textit{Cactus} scene and 10 for the \textit{RootTrans} scene).
We cannot fairly consider the estimate to ground truth metric since NR-NeRF produces surface crossings throughout the volume, thus we crop the total volume to a slightly (10\%) expanded bounding box of the ground truth.
\subsection{D-NeRF \cite{pumarola2020dnerf}} \label{sec:dnerf}
We run D-NeRF~\cite{pumarola2020dnerf} in the manner shown in their code\footnote{\url{https://github.com/albertpumarola/D-NeRF}}.
We use the default parameters provided by the authors and sweep the threshold to find the threshold giving minimal metric (400 for the \textit{Cactus} scene and 400 for the \textit{RootTrans} scene).
Note that D-NeRF has failed to produce reasonable geometry for the \textit{RootTrans} scene, which we hypothesize is due to the large camera motion.
We cannot fairly consider the estimate to ground truth metric for the same reason discussed for D-NeRF in Section~\ref{sec:dnerf}.
The same region cropping to a slightly (10\%) expanded bounding box of the ground truth.
\par
Note that it appears that D-NeRF~\cite{pumarola2020dnerf} has failed for the \textit{RootTrans} scene.
We also try different thresholds (specifically, 40 which is the default in the D-NeRF code) and run the scene 6 times: 3 of these produce no geometry at all for all tried thresholds while 3 produce some geometry not accurately representing the \textit{RootTrans} scene.
\subsection{LASR \cite{yang2021lasr}} \label{sec:lasr}
We run LASR~\cite{yang2021lasr} in the manner shown in their code\footnote{\url{https://github.com/google/lasr}}.
We progressively increase the number of bones and faces in a coarse-to-fine manner following the configurations provided.
This progression is shown in Table~\ref{tab:lasr}.
For the \textit{RootTrans} sequence, we use a slightly modified version of the code.
This was to prevent a complete failure case where bone re-initialization without CNN re-initialization results in the mesh entering a local minimum that no longer reprojects on the image.
This code modification was made with the assistance of the lead author of LASR~\cite{yang2021lasr}.
{\renewcommand{\arraystretch}{1.1}
\begin{table}[!ht]
	\begin{center}
		\begin{tabular}{|c|c|c|c|c|}
			\hline
			\textit{Step} & \textit{Bones} & \textit{Faces} & \textit{Hypotheses} & \textit{Epochs} \\
			\hline
			r1 & 21 & 1280 & 16 & 20 \\
			r2 & 26 & 1600 & 1 & 10 \\
			r3 & 31 & 1920 & 1 & 10 \\
			r4 & 31 & 2240 & 1 & 10 \\
			r5 & 36 & 2560 & 1 & 10 \\
			final & 36 & 2880 & 1 & 10 \\
			\hline
		\end{tabular}
	\end{center}
	\caption
	{
	    A subset of parameters used when running LASR~\cite{yang2021lasr}.
	}
	\label{tab:lasr}
\end{table}
}
\subsection{ViSER \cite{yang2021viser}} \label{sec:viser}
We run ViSER \cite{yang2021viser} in the manner shown in their code\footnote{\url{https://github.com/gengshan-y/viser-release}}.
For the initialization phase, we use the frames ranges shown in Table~\ref{tab:viser}.
The selection of these frame ranges was done in accordance with the guidance that "the viewpoint coverage is large enough" \cite{yang2021viser}.
\begin{table}[!ht]
    \begin{center}
        \begin{tabular}{|c|c|c|}
            \hline
            \textit{Dataset} & \textit{Start frame} & \textit{End frame} \\
            \hline
            Cactus & 10 & 40 \\
            RootTrans & 10 & 40 \\
            \hline
        \end{tabular}
    \end{center}
    \caption
    {
        Frame ranges used for initialization when running ViSER~\cite{yang2021viser}.
    }
    \label{tab:viser}
\end{table}
\subsection{Direct, Dense, Deformable \cite{yu2015direct}} \label{sec:ddd}
We run the method of Yu~et~al.~\cite{yu2015direct} in the manner shown in their code\footnote{\url{https://github.com/cvfish/PangaeaTracking}}.
We empirically explored a set of values and found those of Table~\ref{tab:ddd} to perform best when comparing results after rigid alignment with ICP~\cite{besl1992icp} to the ground truth.
{\renewcommand{\arraystretch}{1.1}
\begin{table}[!ht]
	\begin{center}
		\begin{tabular}{|c|c|}
			\hline
			\textit{Parameter} & \textit{Value} \\
			\hline
			Photometric weight & 1 \\
			ARAP weight & 20 \\
			\hline
		\end{tabular}
	\end{center}
	\caption
	{
	    A subset of parameters used when running the method of Yu~et~al.~\cite{yu2015direct}. If not mentioned otherwise, we use the parameters as proposed in the original code.
	}
	\label{tab:ddd}
\end{table}
}
\subsection{Neural \gls{nrsfm} \cite{sidhu2020neural}} \label{sec:nnrsfm}
We run Neural \gls{nrsfm} in the manner shown in their code\footnote{\url{http://vcai.mpi-inf.mpg.de/projects/Neural_NRSfM/}}.
In order to acquire the \gls{mfof} $W$ matrix used as input by this implementation, we use the Matlab~\cite{MATLAB:2010} code of Ansari~et~al.~\cite{Ansari2017}.
Additionally, this implementation requires input in a specific format which is computed using proprietary code provided by the authors.
The loss function weights used are given in Table~\ref{tab:nnrsfm}.
{\renewcommand{\arraystretch}{1.1}
\begin{table}[!ht]
	\begin{center}
		\begin{tabular}{|c|c|}
			\hline
			\textit{Parameter} & \textit{Value} \\
			\hline
			$\beta$ & $1$ \\
			$\gamma$ & $1\e{-4}$ \\
			$\eta$ & $1$ \\
			$\lambda$ & $0$ \\
			\hline
		\end{tabular}
	\end{center}
	\caption
	{
	    A subset of parameters used when running the method of \cite{sidhu2020neural}.
	}
	\label{tab:nnrsfm}
\end{table}
}
%
%
\subsection{Human Geometry Proxy} \label{sec:hgp}
To generate the proxy geometry for our human character sequences (\textit{i.e.} the \textit{RootTrans} synthetic sequence and the \textit{Humanoid} real-world sequence), we employ SMPLify-X~\cite{SMPL-X:2019} to obtain the root orientation and joint angles of SMPLX~\cite{SMPL-X:2019} human mesh model from the input image sequence. We then solve the 2D reprojection based optimization $\mathcal{L}_{\text{2D}}$ to obtain the 3D root translation of the human mesh:   
\begin{equation}
\mathcal{L}_{\text{2D}}=\frac{1}{K}\sum^{K}_{k=1}  \left \|\Pi (X_{k}) - p_{k}  \right \|_2^{2},
\end{equation}
where $\Pi(\cdot)$ and $K$ represents the perspective projection operator and the number of joints, respectively. $X_{k}$ and $p_{k}$ denotes the $k$th 3D joint keypoint obtained from \cite{SMPL-X:2019}, and pseudo GT 2D joint keypoints obtained from OpenPose \cite{openpose4}. To solve the optimization, we use Adam \cite{adam} optimizer with the camera intrinsics estimated by COLMAP \cite{schoenberger2016sfm}, \cite{schoenberger2016mvs}, and use a fixed height for the human model of 180~cm. Finally, we transform the vertices using the estimated camera extrinsics to place the model in world space.
%
%
\section{Geometric Proxy Resolution Ablation} \label{sec:proxy_ablation}
While our experiments have used a complete SMPLX model~\cite{SMPL-X:2019} for the \textit{RootTrans} scene as described in Section~\ref{sec:hgp}, we have identified that the geometric proxy could be reduced further.
This is shown by the use of only a 12 vertex skeleton for the \textit{Humanoid} scene.
A proxy that summarizes the motion of the scene by coarsely tracking the extremities would be sufficient.
This could allow the use of skeleton tracking systems (e.g.~\cite{VNect_SIGGRAPH2017}) rather than methods with dense mesh output like SMPLify-X~\cite{SMPL-X:2019}.
We validate this possible approach in Figure~\ref{fig:proxy_ablation} by reducing the SMPLX mesh to just 7 vertices (one on each extremity, two on the body, and one on the head).
This 7 vertex proxy is sufficient to constrain Ub4D to produce a single canonical copy, rather than the multiple copies when no proxy is supplied (see Figure~5(b) in the main paper).
\begin{figure}[ht!]
	\begin{center}
		\includegraphics[width=\linewidth]{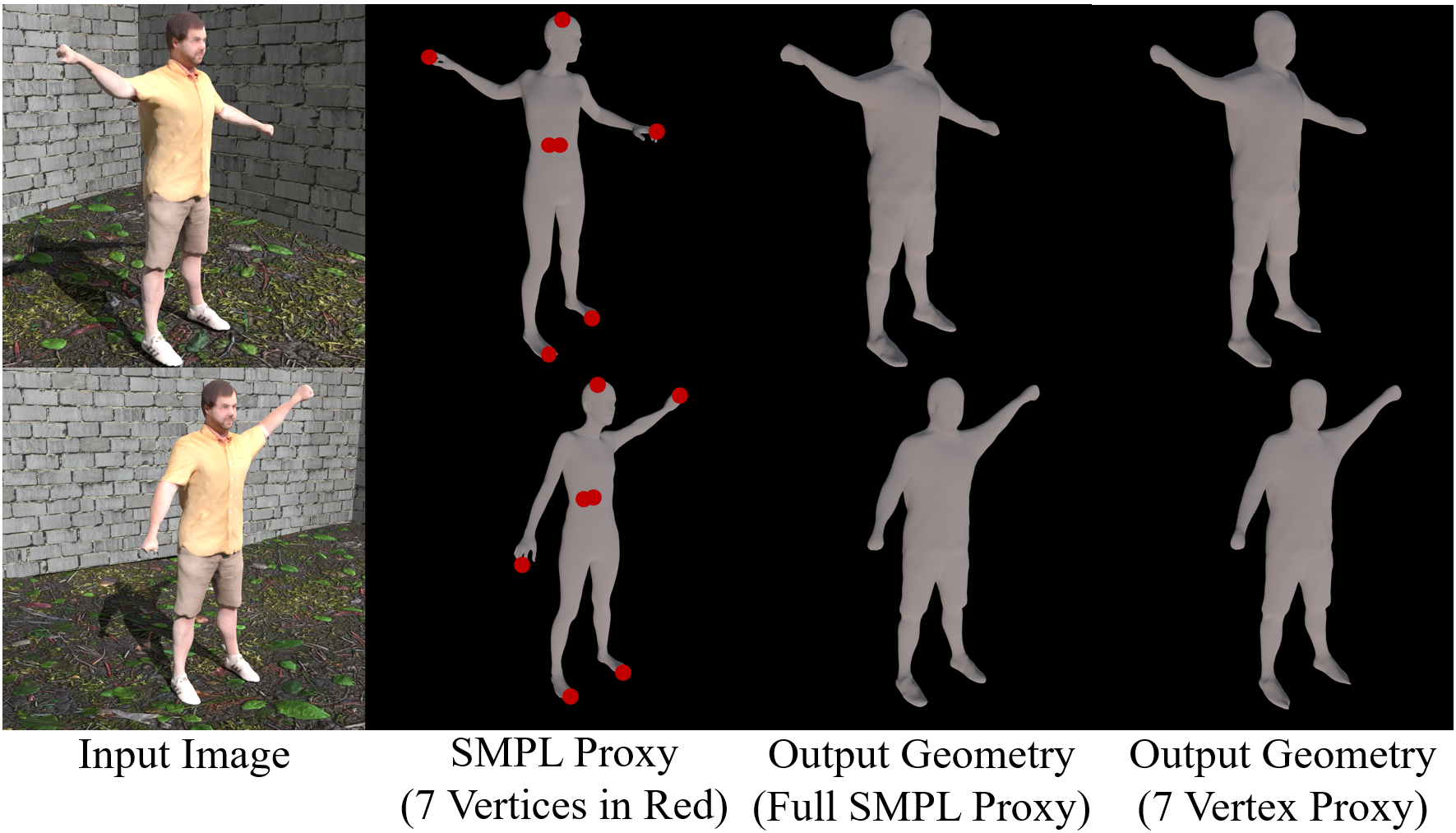}
	\end{center}
	\caption
	{
		Comparison of the scene flow loss with full SMPL proxy ($10.4k$ vertices) and \textit{just seven vertices} from the SMPL proxy. 
	}
	\label{fig:proxy_ablation}
\end{figure}
%
%
\section{Per-Frame Latent Code Analysis and Novel Geometry Synthesis} \label{sec:latents}
In Ub4D, the entirety of the model's understanding of time is encoded into a per-frame latent code provided to the bending network.
Initializing these latent codes with zeros gives our latent space a valuable property: a smooth, semantically meaningful latent representation.
Demonstrating such a latent representation allows us to interpolate latent codes for certain applications, e.g. temporal super-resolution.
It also opens the door for employing such deformation models using latent codes to analyze motion (e.g. periodicity detection, metrically comparing deformation states).
\par
To validate the semantic meaning of our latent representation we perform PCA~\cite{pearson1901pca} on the 64 dimensional learned latent codes.
The results are shown in Figure~\ref{fig:pca_latents}.
Note that even though the latent space is never directly constrained in Ub4D, neighbouring frames (i.e. similar colors in Figure~\ref{fig:pca_latents}) tend to be nearby.
\begin{figure}[ht!]
	\begin{center}
		\includegraphics[width=\linewidth]{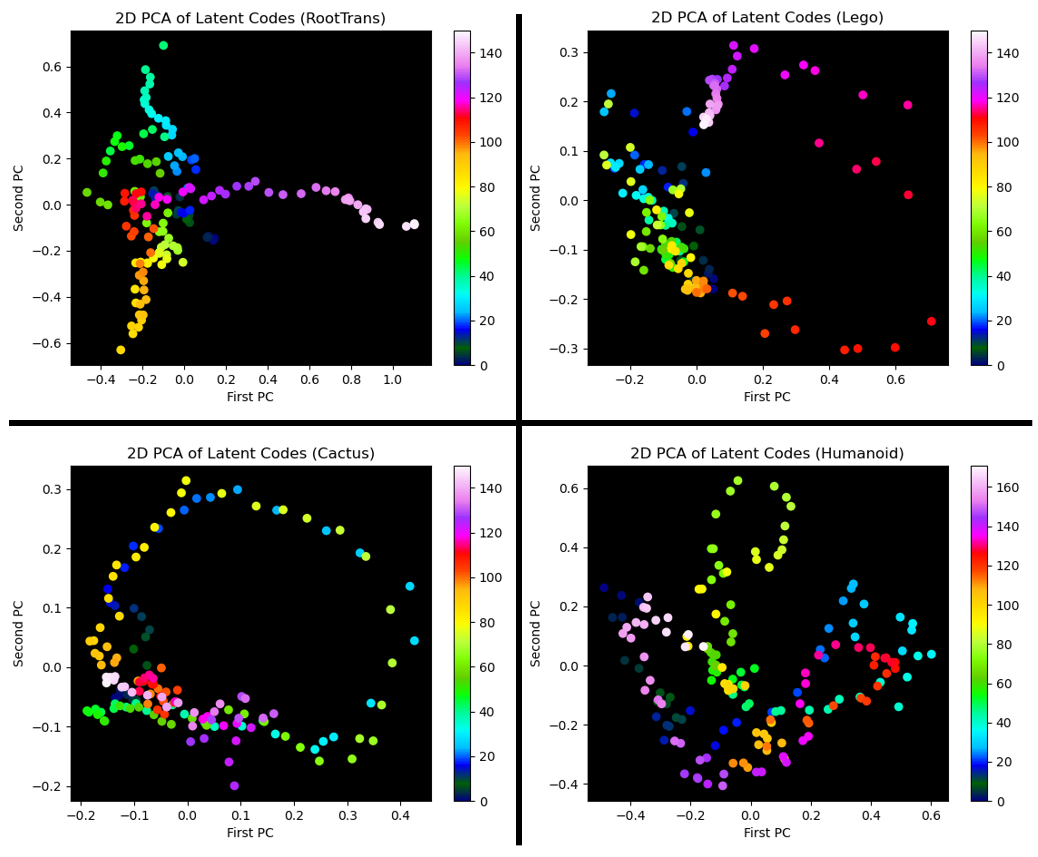}
	\end{center}
	\caption[2D PCA of learned 64D latent codes.]
	{
		2D PCA of learned 64D latent codes for the \textit{RootTrans}, \textit{Cactus}, \textit{Lego}, and \textit{Humanoid} scenes.
		The first two principal components explain $38$\%, $32$\%, $25$\%, and $24$\% of the variance, respectively.
		The colors correspond to frames.
		Note how similar colors are nearby.
	}
	\label{fig:pca_latents}
\end{figure}
\par
We wish to compare against the standard latent code initialization approach: random Gaussian initialization.
However, the same concept of performing PCA~\cite{pearson1901pca} does not suffice.
This is because PCA uses directions of maximum variance and randomly initialized latent codes could structure themselves ``inside'' of the variance.
While a more complex dimensional reduction technique (e.g. t-SNE~\cite{van2008tsne}) could yield results, a failure to visualize a meaningful structure would not definitively show that such a structure does not exist.
Therefore, we use a reduced latent code dimension allowing visualization without dimensional projection.
\par
Taking the \textit{Cactus} scene, we train using 2D latent codes: once initializing with zeroes as proposed in Tretschk~et~al.~\cite{Tretschk2020} and once initializing with random Gaussian samples.
We show the resulting learned latent codes in Figure~\ref{fig:2d_latents}.
Note how spatially coherent the zero-initialized latent codes become during training, whereas the random Gaussian initialized latent codes do not have this property.
Observing the particular structure of the zero-initialized case and slight clustering of similar frames in the random Gaussian initialization, one could imagine these latent codes as charged molecules, with similar states attracting and differing states repelling, resulting in a particular fold.
\begin{figure}[ht!]
	\begin{center}
		\includegraphics[width=\linewidth]{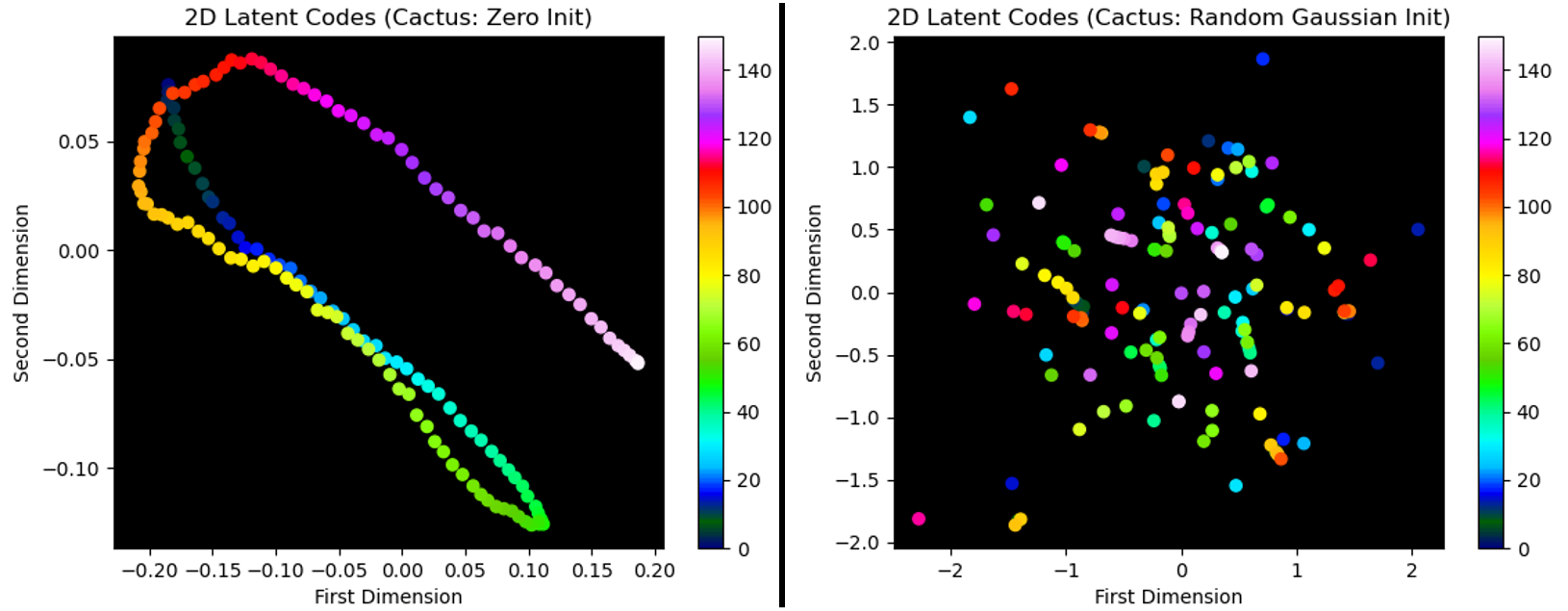}
	\end{center}
	\caption[Learned 2D latent codes for \textit{Cactus} scene.]
	{
		Learned 2D latent codes for the \textit{Cactus} scene showing the latent code provided to the network without any dimensional projection.
		We initialize with zeroes on the left and random Gaussian samples on the right.
		The colors correspond to frames.
		Note how similar colors are nearby for zero-initialized latent codes, whereas the random Gaussian initialization does not give rise to such a property (although some local structuring is interesting).
	}
	\label{fig:2d_latents}
\end{figure}
\par
Some applications require semantically meaningful latent codes which we have demonstrated in the analysis above.
This allows us to generate entirely new geometries by providing novel latent codes.
Figure~\ref{fig:novel_latents} shows samples of a novel latent path for the 2D latent codes from the left-hand side of Figure~\ref{fig:2d_latents} (see webpage or video for a better visualization).
A further investigation is required into the ability to generate new geometries from novel latent codes, particularly when using higher dimensional latent codes or exceeding the convex hull of the observations.
\begin{figure}[ht!]
	\begin{center}
		\includegraphics[width=\linewidth]{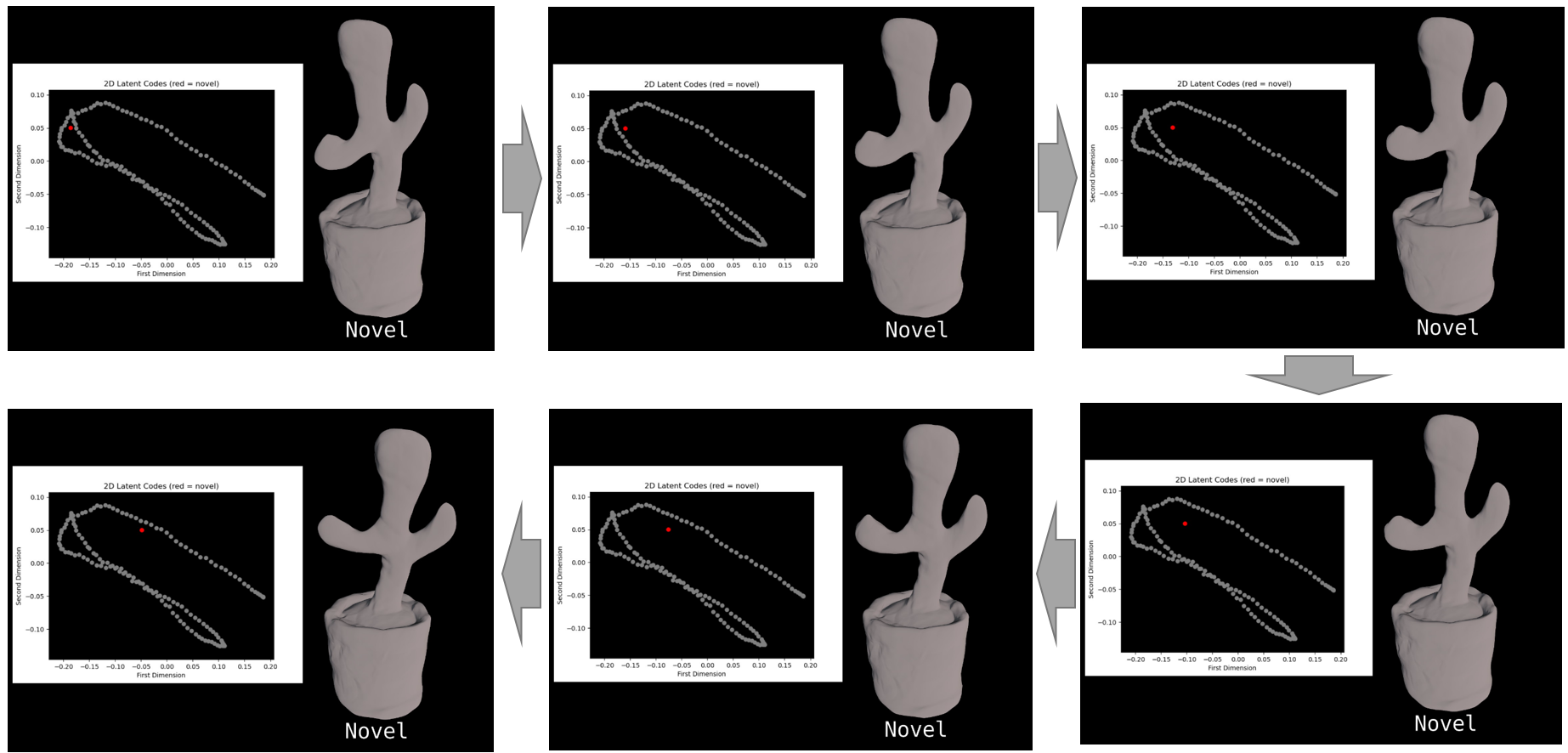}
	\end{center}
	\caption[Synthesizing new geometries with novel latent codes.]
	{
	    Synthesizing entirely new geometries with novel latent codes.
	    Red dot in latent space plot shows the provided latent code while grey dots show the original sequence.
	    Note the smoothness and plausibility of the deformation.
	}
	\label{fig:novel_latents}
\end{figure}
%
%
\section{Additional Limitations} \label{sec:additional_limitations}
One limitation of Ub4D is that errors in the geometric proxy can increase the Chamfer distance of our reconstruction.
Our scene flow loss is designed such that it does not require highly accurate correspondences to allow us to handle large deformations and prevent multiple canonical copies; however, we still inherit errors from the geometric proxy.
Figure \ref{fig:roottrans_failure} illustrates this for a geometric proxy that is offset from the ground truth which results in the region with a high error on our reconstruction.
Note that our method results in a decreased Chamfer distance for this frame compared to the geometric proxy (0.76 vs 0.92).
\par
Another limitation is that acquiring a geometric proxy may limit application if results without the scene flow loss are not satisfactory.
Figure \ref{fig:supp_humanoid_limitations} shows one example of a common issue encountered without the scene flow loss.
In this case, an additional appendage is used to satisfy the reconstruction losses while not grossly violating the other regularizers of our method.
\begin{figure}[ht!]
	\begin{center}
		\includegraphics[width=0.88\linewidth]{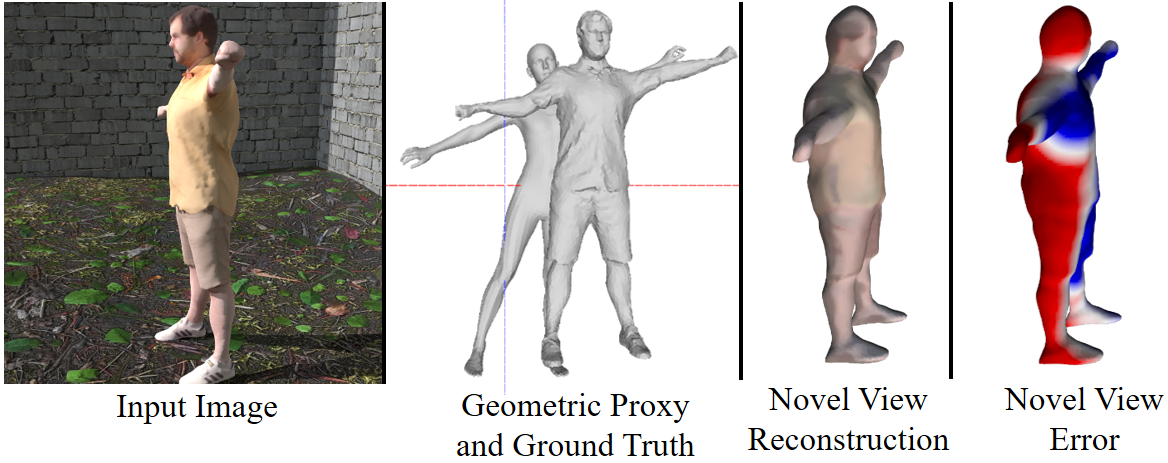}
	\end{center}
	\caption
	{
	    Impact of a significant geometric proxy error. Red regions have a higher Chamfer distance to the ground truth.
	}
	\label{fig:roottrans_failure}
\end{figure}
\begin{figure}[ht!]
	\begin{center}
		\includegraphics[width=0.75\linewidth]{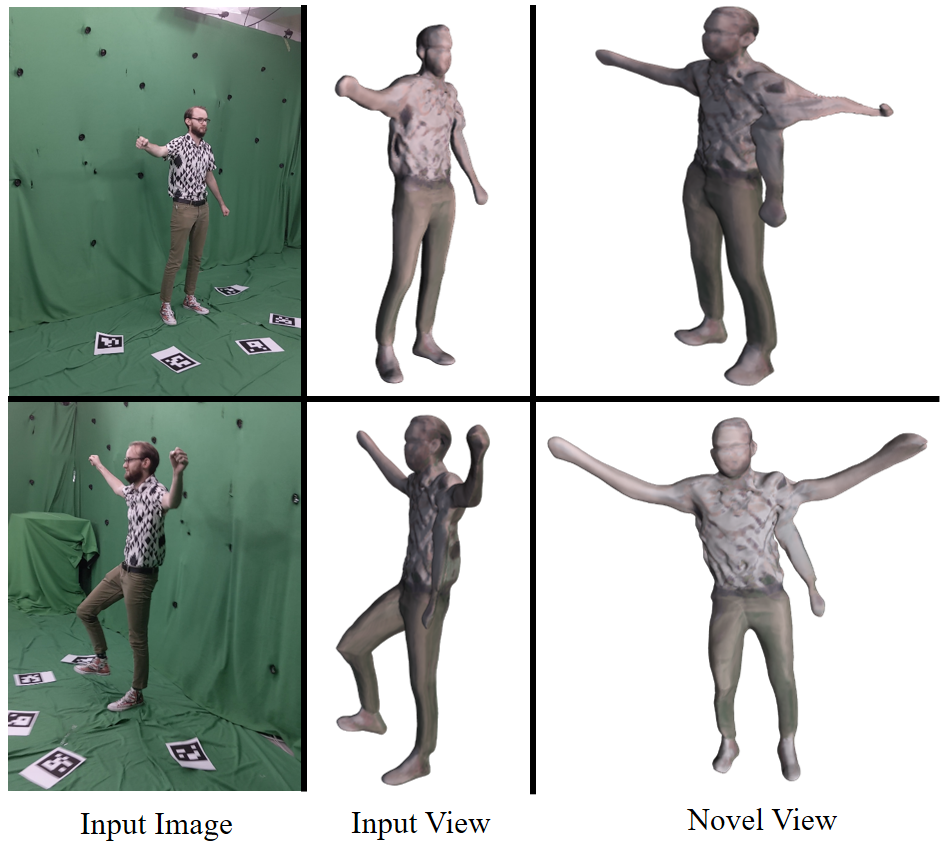}
	\end{center}
	\caption
	{
	    Example of our method without the scene flow loss using an additional appendage to satisfy the reconstruction losses.
	}
	\label{fig:supp_humanoid_limitations}
\end{figure}

\clearpage

\end{document}